\documentclass[10pt,twocolumn,letterpaper]{article}
\usepackage[accsupp]{axessibility}
\usepackage{wacv}
\usepackage{times}
\usepackage{epsfig}
\usepackage{graphicx}
\usepackage{amsmath}
\usepackage{amssymb}
\usepackage{booktabs}
\usepackage{graphicx}
\usepackage[utf8]{inputenc} 
\usepackage[T1]{fontenc}    
\usepackage{url}            
\usepackage{booktabs}       
\usepackage{amsfonts}       
\usepackage{nicefrac}       
\usepackage{microtype}      
\usepackage{xcolor}         

\usepackage{graphicx}
\usepackage{multirow}
\usepackage{makecell}
\usepackage{enumitem}
\usepackage{algorithm,algpseudocode}
\usepackage{mathtools}

\definecolor{LightCyan}{rgb}{0.88,1,1}
\definecolor{LightRed}{rgb}{1.0, 0.91, 0.91}
\definecolor{LightGray}{rgb}{0.88,0.88,0.88}
\definecolor{VeryLightGray}{rgb}{0.93,0.93,0.93}

\makeatletter

\DeclareMathOperator{\E}{\mathbb{E}}
\DeclareMathOperator{\R}{\mathbb{R}}
\DeclareMathOperator{\xx}{\mathrm{x}}
\DeclareMathOperator{\yy}{\mathrm{y}}

\algnewcommand{\algorithmicforeach}{\textbf{for each}}
\algdef{SE}[FOR]{ForEach}{EndForEach}[1]
  {\algorithmicforeach\ #1\ \algorithmicdo}
  {\algorithmicend\ \algorithmicforeach}

\wacvfinalcopy
\usepackage[pagebackref=true,breaklinks=true,letterpaper=true,colorlinks,bookmarks=false]{hyperref}
\hypersetup{linkcolor=blue,anchorcolor=blue,citecolor=cyan,filecolor=blue,urlcolor=blue}
\begin{document}

\title{Improving Diversity with Adversarially Learned Transformations \\for Domain Generalization} 

\author{
Tejas Gokhale\thanks{Work done during internship at LLNL} $~{}^\S$ ~~ Rushil Anirudh $^\#$ ~~ Jayaraman J. Thiagarajan $^\#$ ~~ Bhavya Kailkhura $^\#$ \\ Chitta Baral $^\S$ ~~ Yezhou Yang $^\S$\\
$~{}^\S$ Arizona State University ~~ $^\#$ Lawrence Livermore National Laboratory \\
{\tt\small \{tgokhale, chitta, yz.yang\}@asu.edu ~~ \{anirudh1, jjayaram, kailkhura1\}@llnl.gov}
}

\maketitle
\thispagestyle{empty}

\begin{abstract}
To be successful in single source domain generalization (SSDG), maximizing diversity of synthesized domains has emerged as one of the most effective strategies. 
Recent success in SSDG comes from methods that pre-specify diversity inducing image augmentations during training, so that it may lead to better generalization on new domains.
However, na\"ive pre-specified augmentations are not always effective, either because they cannot model large domain shift, or because the specific choice of transforms may not cover the types of shift commonly occurring in domain generalization. 
To address this issue, we present a novel framework called ALT: adversarially learned transformations, that uses an adversary neural network to model plausible, yet hard image transformations that fool the classifier.
ALT learns image transformations by randomly initializing the adversary network for each batch and optimizing it for a fixed number of steps to maximize classification error.
The classifier is trained by enforcing a consistency between its predictions on the clean and transformed images.
With extensive empirical analysis, we find that this new form of adversarial transformations achieves both objectives of diversity and hardness simultaneously, outperforming all existing techniques on competitive benchmarks for SSDG. 
We also show that ALT can seamlessly work with existing diversity modules to produce highly distinct, and large transformations of the source domain leading to state-of-the-art performance.
Code: {\small\url{https://github.com/tejas-gokhale/ALT}}
\end{abstract}

\section{Introduction}
\begin{figure*}[t]
    \centering
    \includegraphics[width=0.86\linewidth]{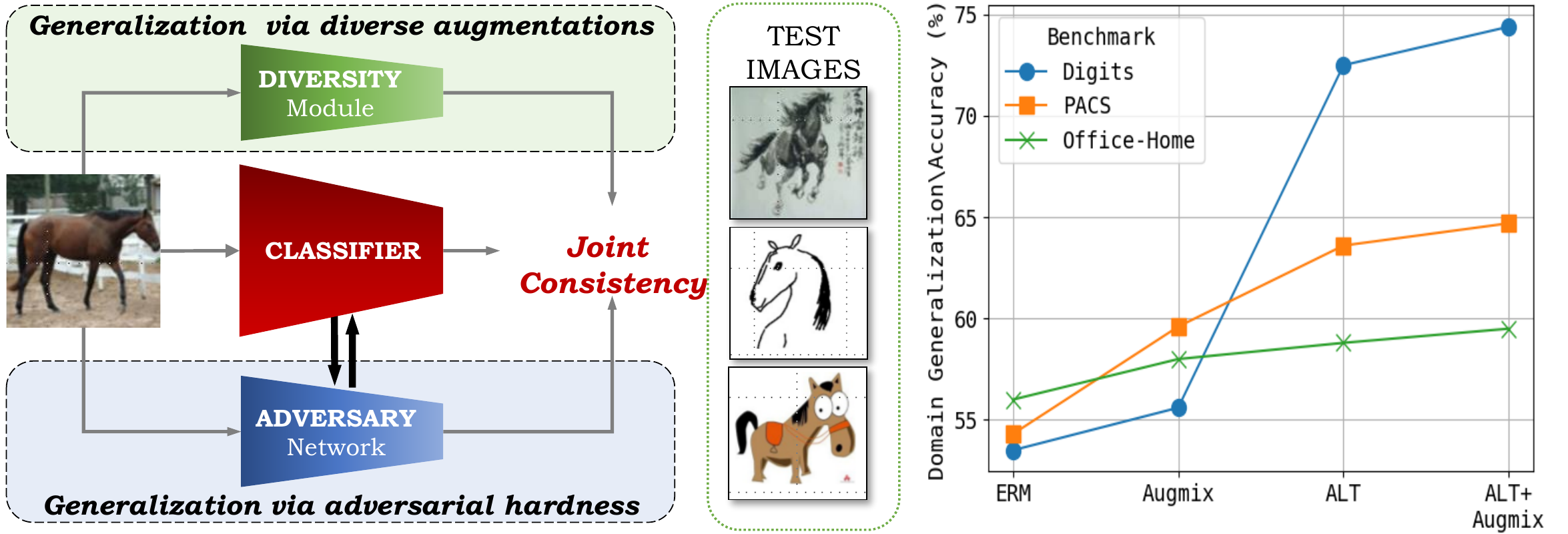}
    \caption{
    ALT consists of a \textit{diversity} module (data augmentation functions such as Augmix~\cite{hendrycks2019augmix} or RandConv~\cite{xu2020robust} and an \textit{adversary} network (to learn image transformations that fool the classifier).
    We show an example from the PACS benchmark under the single-source domain generalization setting, with real photos (P) as the source domain and art paintings (A), cartoons (C), and sketches (S) as the target domains.
    The plot summarizes our results -- while diversity alone improves performance over the naive ERM baseline, adapting this diversity using adversarially learned transformations (ALT) provides a significant boost for domain generalization on multiple benchmarks.
    }
    \label{fig:alt_teaser}
\end{figure*}

Domain generalization is the problem of making accurate predictions on previously unseen domains, especially when these domains are very different from the data distribution on which the model was trained. 
This is a challenging problem that has seen steady progress over the last few years~\cite{carlucci2019domain,volpi2018generalizing,qiao2020learning,xu2020robust,nam2021reducing}. 
This paper focuses on the special case -- single source domain generalization (SSDG) -- where the model has access only to a single training domain, and is expected to generalize to multiple different testing domains. 
This is especially hard because of the limited information available to train the model with just a single source. 

When multiple source domains are available (MSDG), recent analysis~\cite{gulrajani2021in} shows that even simple methods like minimizing empirical risk jointly on all domains, performs better than most existing sophisticated formulations.
A corollary to this finding is that success in DG is dependent on \emph{diversity} -- i.e., exposing the model to as many potential training domains as possible.
As the SSDG problem allows access only to a single training domain, such an exposure must come in the form of diverse transformations of the source domain that can simulate the presence of multiple domains, ultimately leading to low generalization error. 

The idea of using diversity to train models has been sufficiently explored -- \cite{hendrycks2019augmix,yun2019cutmix,zhang2018mixup,cubuk2020randaugment} show that a diverse set of augmentations during training improves a model's robustness under distribution shifts. 
Specific augmentations can be used if the type of diversity encountered at test time is known; for eg., if it is known that the test set contains random combinations of rotation, translation, and scaling, using augmentations correlated with this domain shift would lead to good performance~\cite{benton2020learning,wong2020learning,gokhale2021attribute}.
However, since we cannot assume knowledge of the test domain under the SSDG problem statement, the extent to which the model needs to be exposed to specific augmentations remains unclear.
Augmentation methods impose a strong prior in terms of the types of diversity that the model is exposed to, which may not match with desirable test-time transformations.
As we will show in this paper, data augmentation methods that produce good results on one dataset, do not necessarily work on other datasets -- in some cases, they may even hurt performance!

In addition to such a knowledge gap, unfortunately, such augmentation methods can only achieve invariance under small distribution shifts like unknown corruptions, noise, or adversarial perturbations, but do not work effectively when the distribution shift is large and of a semantic nature, as in the case of domain generalization. 
On the other hand, some recent methods have directly used randomized convolutions to synthesize diverse image manipulations~\cite{xu2020robust}, motivated by the large space of potentially realizable functions induced by a convolutional layer, which cannot be easily emulated using simple analytical functions.

In this paper we hypothesize that, while diversity is necessary for single-source domain generalization, diversity alone is insufficient -- blindly exposing a model to a wide range of transformations may not guarantee greater generalization.
Instead, we argue that carefully designed forms of diversity are needed -- specifically those that can expose the model to unique and task-dependent transformations with large semantic changes that are otherwise unrealizable with plug-and-play augmentations as before. 
To this end, we introduce an adversary network whose objective is to \textit{find} plausible image transformations that maximize classification error. 
This adversary network enables access to a much richer family of image transformations as compared to prior work on data augmentation.
By randomly initializing the adversary network in each iteration, we ensure the adversarial transformations are unique and diverse themselves.
We enforce a consistency between a \textbf{diversity module} and the \textbf{adversary network} during training along with the classifier's predictions, so that together they expose the model to learn from both diverse and challenging domains.

Our method, dubbed ALT (adversarially learned transformations), offers an interplay between diversity and adversity. 
Over time, a synergistic partnership between the diversity and adversary networks emerges, exposing the model to increasingly unique, challenging and semantically diverse examples that are ideally suited for single source domain generalization.  
The adversary network benefits from the classifier being exposed to the diversity module, and as such avoid trivial adversarial samples with appropriate checks. 
This allows the adversarial maximization to explore a wider space of adversarial transformations that cannot be covered by prior work on pixel-level additive perturbations.
  
We demonstrate this advantage of our method empirically on multiple benchmarks -- PACS~\cite{li2017deeper}, Office-Home~\cite{venkateswara2017deep}, and Digits~\cite{volpi2018generalizing}. 
On each benchmark, we outperform the state-of-the-art single source domain generalization methods by a significant margin.  Moreover, since our framework disentangles diversity and adversarial modules, we can combine it with various diversity enforcing techniques -- we identify two such state-of-the-art methods with AugMix~\cite{hendrycks2019augmix}, and RandConv~\cite{xu2020robust}, and show that placing them inside our framework leads to significantly improved generalization performance over their vanilla counterparts. 
We illustrate this idea in Figure \ref{fig:alt_teaser} where we show an image of a horse from the `photo' training distribution in PACS and the different styles of cartoon/sketch/art painting horses that may be encountered at test time. 

\noindent\textbf{Contributions:} We summarize our contributions below.
\begin{itemize}[nosep,noitemsep,leftmargin=*]
    \item We introduce a method, dubbed ALT, which produces adversarially learned image transformations that expose a classifier to a large space of image transformations for superior domain generalization performance. ALT performs adversarial training in the parameter space of an adversary network as opposed to pixel-level adversarial training.
    \item We show how ALT integrates diversity-inducing data augmentation and hardness-inducing adversarial training in a synergistic pipeline, leading to diverse transformations that cannot be realized by blind augmentation strategies or adversarial training methods on their own.
    \item We validate our methods empirically on three benchmarks (PACS, Office-Home, and Digits) demonstrating state-of-the-art performance and provide analysis of our approach.
\end{itemize}

\section{Related Work}
\textbf{Multi-Source Domain Generalization.}
Domain generalization has been explored under both multi-source (MSDG) and single-source (SSDG) settings. 
For the MSDG task, multiple source domains are available for training and performance is evaluated on other unseen target domains.
Techniques designed for MSDG seek to utilize these multiple domains to perform feature fusion~\cite{shen2019situational}, learning domain-invariant features~\cite{ganin2016domain}, meta-learning~\cite{li2018learning}, invariant risk minimization~\cite{arjovsky2019invariant}, learning mappings between multiple training domains~\cite{robey2021model}, style randomization~\cite{nam2021reducing}, and learning a conditional generator to synthesize novel domains using cycle-consistency~\cite{zhou2020learning} 
Gulrajani \textit{et al.}~\cite{gulrajani2021in} provide an extensive comparative study of these approaches and report that simply performing ERM on the combination of source domains leads to the best performance. 
Many benchmarks have been proposed to evaluate MSDG performance such as PACS~\cite{li2017deeper}, OfficeHome~\cite{venkateswara2017deep}, Digits~\cite{volpi2018generalizing}, and WILDS~\cite{koh2021wilds} which is a compendium of MSDG datasets.

In the \textbf{Single-Source Domain Generalization} setting, only one domain is available for training, and as SSDG is harder as MSDG methods are infeasible; most work has therefore focused on data augmentation.
Notable among these methods is the idea of adversarial data augmentation -- ADA\cite{volpi2018generalizing} and M-ADA~\cite{qiao2020learning} apply pixel-level additive perturbations to the image in order to fool the classifier.  Resulting images are used as augmented data to train the classifier.
RandConv~\cite{xu2020robust} shows that shape-preserving transformations in the form of random convolutions of images lead to impressive performance gains on Digits.

\textbf{Adversarial Attack and Defense.}
Adversarial attack algorithms have been developed to successfully fool image classifiers via pixelwise perturbations~\cite{goodfellow2014explaining,moosavi2016deepfool,carlini2017towards,dong2018boosting}.
Algorithms have been developed to defend against such adversarial attacks~\cite{moosavi2016deepfool,dhillon2018stochastic,yuan2019adversarial,jang2019adversarial}.
The scope of this paper is not to perform adversarial attack and defense, but to develop a framework to obtain adversarially generated samples that improve domain generalization performance.

\textbf{Adversarial Training.} 
In ALT, we emphasize on the nature of the diversity that could be acquired during training, which is crucial in the single-source setting. 
ALT learns adversarial perturbations in the function space of \emph{neural network weights}.
This allows us access to a wider and richer space of augmentations compared to pixel-wise perturbations such as ADA and M-ADA, or combinatorial augmentation search methods such as ESDA~\cite{volpi2019addressing}.
The adversarial component in ALT allows the network to seek newer and harder transformations for every batch as training progresses, which cannot be achieved with static augmentations such as AugMix or RandConv, or by utilizing normalization layer statistics for style debiasing~\cite{nam2021reducing}.

\textbf{Robustness to Image Corruptions.}
There has also been interest in training classifiers that are robust to corruptions that occur in the real world, such as different types of noise and blur, artifacts due to compression techniques, and weather-related environments such as fog, rain, and snow.
\cite{vasiljevic2016examining,geirhos2018generalisation} show that training models with particular types of corruption augmentations does not guarantee robustness to other unseen types of corruptions or different levels of corruption severity.
Hendrycks~ \textit{et al.}~\cite{hendrycks2018benchmarking} curate benchmarks (ImageNet-C and CIFAR-C) to test robustness along a fixed set of corruptions.
They also provide a benchmark called ImageNet-P which tests robustness against other corruption types such as small tilts and changes in brightness.
A similar benchmark for corruptions of handwritten digit images, MNIST-C~\cite{mu2019mnist} has also been introduced.

\textbf{Data Augmentation}
has been an effective strategy for improving in-domain generalization using simple techniques such as random cropping, horizontal flipping~\cite{he2016deep}, occlusion or removal of patches~\cite{devries2017improved,zhong2020random}.
Data augmentation techniques have been shown to improve robustness against adversarial attacks and natural image corruptions~\cite{zhang2018mixup,yun2019cutmix,cubuk2020randaugment}.
Learning to augment data has been explored in the context of object detection~\cite{zoph2020learning} and image classification~\cite{ratner2017learning,cubuk2019autoaugment,zhang2019adversarial}.

\section{Proposed Approach} 
Under the single-source domain generalization setting, consider the training dataset $\mathcal{D}$ containing $N$ image-label pairs $\mathcal{D} = \{(\xx_i, \yy_i)\}_{i=1}^{N}$, and a classifier $f$ parameterized by neural network weights $\theta$. 
The standard expected risk minimization (ERM) approach seeks to learn $\theta$ by minimizing the in-domain risk measured by a suitable loss function such as the cross-entropy loss.
\begin{equation}
    \label{eq:erm}
    \mathcal{R}_{ERM} = \underset{\xx\in\mathcal{D}}{\E} \mathcal{L}_{CE}(f(\xx;\theta), \yy).
\end{equation}
For SSDG, we are interested in a classifier that has the least risk on several \textit{unseen} target domains $\mathcal{D}^\prime$ that are not observed during training. 
We consider SSDG under covariate shift, i.e. when $P(X)$ changes but $P(Y|X)$ remains the same.
Our approach builds on diversity based and adversarial augmentation approaches which we outline next. 

\medskip
\noindent\textbf{Generalization via Maximizing Diversity.} A successful strategy to improve generalization on unseen domains is to utilize a set of pre-defined data augmentations $\mathcal{F}_{div}$, to emphasize the invariance properties that are important for $f(\theta)$ to learn. 
Such methods modify Equation~\ref{eq:erm} as:
\begin{equation}
    \label{eq:div_erm}
    \mathcal{R}_{div} = \underset{\xx\in\mathcal{D}}{\E} \mathcal{L}_{CE}(f(\xx; \theta), \yy) + \lambda_{KL} D_{KL},
\end{equation}
where $D_{KL}$ is a consistency term, typically a divergence, such as KL-Divergence, between the softmax probabilities of the classifier obtained with the clean and transformed data, respectively, \textit{e.g.}, $D_{KL} = KL(f(\xx) ||f(\mathcal{F}_{div}(\xx)))$. 
The choice of $\mathcal{F}_{div}$ leads to different types of augmentations; for instance, AugMix~\cite{hendrycks2019augmix} utilizes a combination of pre-defined transformations such as shear, rotate, color jitter, 
An approach proposed by Xu \textit{et al.}~\cite{xu2020robust} is to apply a randomly initialized convolutional layer to the input image.
Methods such as these are effective strategies to enforce diversity-based consistencies for generalization.
Although these methods have the advantage of being simple pre-defined transformations that are dataset agnostic, they suffer from drawbacks under the SSDG setting.
When executed on their own, they may not capture sufficient diversity in terms of \textbf{large} semantic shifts, such as when expecting generalization on sketches from a model trained on photos.

\medskip
\noindent\textbf{Generalization via Adversarial Hardness.}
An alternative domain generalization approach is via adversarial augmentation which exposes a classifier to `hard' samples during training -- defined broadly as examples that are carefully designed to cause the model to fail. 
Such samples are augmented to the training set, with the expectation that exposure to such adversarial examples can improve the model's generalization performance on unseen domains \cite{volpi2018generalizing,qiao2020learning}. This is commonly enforced by learning an additive noise vector which when added, maximizes classifier cost. Unfortunately in the case of domain generalization, these methods have failed to match the performance of diversity-only methods optimizing for the cost outlined in Equation~\ref{eq:div_erm}. This is in part because they lack sufficient diversity, and by design they can only guarantee robustness to small perturbations from the training domain, as opposed to large semantic and stylistic shifts, which are crucial for domain generalization.

\subsection{ALT: Adversarially Learned Transformations}
While diversity-only methods have shown promise, they are limited in their ability to generalize to domains with large shifts. 
On the  other hand, techniques based purely on adversarial hardness are theoretically well-motivated but do not match the performance of diversity-based methods. 
In this paper, we propose a new approach that takes the best of these two approaches using an adversary network that is trained to create \textit{semantically consistent}
image transformations that fool the classifier. 
These manipulated images are then used during training as examples on which the image must learn invariance. Since these perturbations are parameterized as learnable weights of a neural network, the network is free to choose large, complex transformations without being restricted to additive noise as done in previous work~\cite{volpi2018generalizing}. Further, this network is randomly initialized for each batch, making the types of adversarial transformations discovered unique and diverse over the course of training.  
Formally, the adversary network $g$ transforms the input image as 
\begin{equation}
    \xx_g = g(\xx), \textit{~~where~~}
    g\colon\R^{C\times H\times W}\rightarrow\R^{C\times H\times W}
\end{equation}
where $C$, $H$, $W$ are the number of channels, height, and width of input images.
$g$ is parameterized by weights $\phi$. This network, dubbed ALT, forms the backbone of our method.  

\begin{algorithm}[t]
    {\footnotesize
    \caption{Adaptive Diversity via ALT}
    \begin{algorithmic}[0]
        \State \textbf{Input:} Source dataset $\mathcal{D}=\{(\xx_i,\yy_i)\}_{i=1}^N$ 
        \State \textbf{Output:} Network Parameters $\theta^*$  
    \end{algorithmic}
    \begin{algorithmic}[1]
        \State \textbf{Initialize:} $\theta\gets\theta_0$ \algorithmiccomment{weights of $f()$}
        \ForEach{$t \in \{1\dots T\}$}%
            \State $\xx_t, \yy_t \sim\mathcal{D}$ \algorithmiccomment{\textit{sample input batch}}
            \If{$t < T_{pre}$} 
                \State $\theta\gets\theta - \eta\nabla \mathcal{L}_{CE}(f(\xx_t;\theta), \yy_t))$ 
            \Else
                \State $\rho\gets\rho_0,~\phi\gets\phi_0$\algorithmiccomment{weights of $r()$, $g()$}
                \ForEach{$i\in{1\dots m_{adv}}$}
                    \State $\hat{\yy_g} \gets f(g(\xx;\phi);\theta)$
                    \State $\phi\gets\phi+\nabla(\mathcal{L}_{cls}(\hat{\yy_g}, \yy) - \mathcal{L}_{TV}(\xx_g))$
                \EndForEach
                \State $\theta\gets\theta-\eta_{adv}\nabla\mathcal{L}_{ALT}$ \algorithmiccomment{see Equation \ref{eq:L_KL}, \ref{eq:L_ALT}}
            \EndIf 
        \EndForEach
        \State\Return $\theta$
\end{algorithmic}
\label{algo}
}
\end{algorithm}

To train ALT, we setup an adversarial optimization problem with the goal of producing transformations, which when applied to the source domain, can fool the classifier $f$.
While existing efforts dealing with robustness to small corruptions use $\ell_p$ norm-bounded pixel-level perturbations to fool the model, we find that this is not sufficient for domain generalization as such methods do not allow searching for adversarial samples with semantic changes.
Instead, we directly perform adversarial training in the space of $\phi$, i.e., the neural network weights of ALT. Given input images $\xx$, parameters $\phi$ are randomly initialized, and the corresponding adversarial samples $\xx_g$ are found as:
\begin{equation}
    \xx_g = \underset{\phi}{\textrm{max}}~\mathcal{L}_{CE}(f(g(\xx;\phi); \theta), \yy) - \mathcal{L}_{TV}(g(\xx;\phi)). 
    \label{eq:adv_max}
\end{equation}
The first term seeks to update $\phi$ to maximize the classifier loss, while $\mathcal{L}_{TV}$ (total variation)~\cite{rudin1992nonlinear} acts as a smoothness regularization for the generated image $\xx_g = g(\xx;\phi)$. 
The maximization in Eq.~\ref{eq:adv_max} is solved by performing $m_{adv}$ steps of gradient descent with learning rate $\eta_{adv}$. 
We note a few important aspects of ALT -- unlike existing methods that explicitly place an $\ell_p-$norm constraint on the adversarial perturbations, we control the strength of the adversarial examples by limiting the number of optimization steps taken by $g$ to maximize classification error. 
Next, since we randomly initialize $g$ for each batch, the network is reset to a random function. 
In fact, when the number of adversarial steps is set to 0, $g$ behaves similar to RandConv \cite{xu2020robust} since it is only a set of convolutional layers, with additional non-linearity. Finally, in addition to limiting the number of adversarial steps, we place a simple total variation loss on the generated image to force smoothness in the output. This naturally suppresses high frequency noise-like artifacts and encourages realistic image transformations.
It also prevents the optimization from resorting to learning trivial transformations in order to maximize classifier loss, such as noise addition or entirely removing or obfuscating the semantic content of the image.

\medskip
\noindent\textbf{Improving Diversity.}
The samples $\xx_g$ obtained by solving Equation~\ref{eq:adv_max} represent hard adversarial images that can be leveraged by the model to generalize to domain shift. But it also lends itself to exploit other forms of na\"ive diversity achieved by methods like RandConv and AugMix.
We represent these ``diversity modules'' as $r$, which produce outputs $\xx_r=r(\xx)$. Our method utilizes these samples in the training process by enforcing a consistency between the predictions of the classifier on the source image and its transformations from $r$ and $g$. By including the diversity module into the optimization process, the invariances inferred by the classifier lead to stronger and more diverse adversarial examples in future epochs. 
Eventually, a synergistic partnership emerges between the diversity module and the adversary network to produce a wide range of image transformations that are significantly different from the source domain.  

Let $p_c$, $p_r$, and $p_g$ denote the softmax prediction probabilities of classifier $f$ on $\xx$, $\xx_r$, and $\xx_g$, respectively. Then the consistency between these predictions can be computed using Kullback-Leibler divergence~\cite{kullback1951information} as:
\begin{align}
    \mathcal{L}_{KL} = D_{KL}(p_{mix}||p_c) &+ w_r D_{KL}(p_{mix}||p_r) \nonumber\\
    &+ (2-w_r) D_{KL}(p_{mix}||p_g), 
    \label{eq:L_KL}
\end{align}
where $p_{mix}$ denotes the mixed prediction:
\begin{equation}
    p_{mix} = \frac{p_c + w_r p_r + (2-w_r)p_g}{3}.
    \label{eq:p_mix}
\end{equation}
The weight $w_r\in[0,2]$ controls the relative contribution of diversity and adversity to the consistency loss; 
$w_r>1$ implies more weight on consistency with the diversity module;
$w_r<1$ implies more weight on consistency with the adversary network.
In our experiments, we use $w_r=1$, i.e., both diversity and adversary are given equal importance.

Our final loss function for training the classifier is given as the convex combination of the consistency $\mathcal{L}_{KL}$ and the classifier loss $\mathcal{L}_{cls} = \mathcal{L}_{CE}(f(g(\xx); \theta), \yy)$, as shown below:
\begin{equation}
    \mathcal{L}_{ALT} = (1-\lambda_{KL})\mathcal{L}_{cls} + \lambda_{KL}\mathcal{L}_{KL}.
    \label{eq:L_ALT}
\end{equation}

\medskip
\noindent\textbf{Implementation.}
Algorithm~\ref{algo} shows how ALT is implemented.
In our experiments, we use RandConv or AugMix as the diversity module $r$ and a fully-convolutional image-to-image network as the adversary network $g$.
$g$ has 5 convolutional layers with kernel size $3$ and LeakyReLU activation.
We train the classifier for a total of $T$ batch iterations of which $T_{pre}$ iterations are used for pre-training the classifier using standard ERM on only the source domain (with only $\mathcal{L}_{cls}$).
During each batch iterations $t>T_{pre}$, we randomly initialize the weights of both $r$ and $g$ with the ``Kaiming Normal'' strategy~\cite{he2016deep} as our starting point for producing diverse perturbations, and update $g$ using the adversarial cost in Equation~\ref{eq:adv_max}.
After $g$ is adversarially updated for the given batch, we use the combination of classifier loss and consistency in Equation~\ref{eq:L_ALT} to update model parameters $\theta$.

\section{Experiments}
\begin{table*}[t]
    \centering
    \small
    \resizebox{0.9\linewidth}{!}{
    \begin{tabular}{@{}l ccc ccc ccc ccc c@{}}
        \toprule
        \textbf{Method} 
        & \textbf{A$\rightarrow$C} & \textbf{A$\rightarrow$S} & \textbf{A$\rightarrow$P} 
        & \textbf{C$\rightarrow$A} & \textbf{C$\rightarrow$S} & \textbf{C$\rightarrow$P} 
        & \textbf{S$\rightarrow$A} & \textbf{S$\rightarrow$C} & \textbf{S$\rightarrow$P} 
        & \textbf{P$\rightarrow$A} & \textbf{P$\rightarrow$C} & \textbf{P$\rightarrow$S} 
        & \textbf{Avg.}\\
        \midrule
        ERM                                 & 62.3 & 49.0 & 95.2 & 65.7 & 60.7 & 83.6 & 28.0 & 54.5 & 35.6 & 64.1 & 23.6 & 29.1 & 54.3\\
        JiGen~\cite{carlucci2019domain}     & 57.0 & 50.0 & 96.1 & 65.3 & 65.9 & 85.5 & 26.6 & 41.1 & 42.8 & 62.4 & 27.2 & 35.5 & 54.6\\
        ADA~\cite{volpi2018generalizing}    & 64.3 & 58.5 & 94.5 & 66.7 & 65.6 & 83.6 & 37.0 & 58.6 & 41.6 & 65.3 & 32.7 & 35.9 & 58.7 \\
        SagNet~\cite{nam2021reducing}       & 67.1 & 56.8 & 95.7 & 72.1 & 69.2 & 85.7 & 41.1 & 62.9 & 46.2 & 69.8 & 35.1 & 40.7 & 61.9 \\
        RandConv~\cite{xu2020robust}        & 61.1 & 60.5 & 87.3 & 57.1 & 72.9 & 73.7 & 52.2 & 63.9 & 46.1 & 61.3 & 37.6 & 50.5 & 60.3\\
        AugMix~\cite{hendrycks2019augmix}   & 68.4 & 54.6 & 95.2 & 74.3 & 66.7 & 87.3 & 40.0 & 57.4 & 46.8 & 67.3 & 26.8 & 41.4 & 59.6 \\
        RandConv{+}AugMix                   & 64.2 & 62.5 & 90.7 & 65.4 & 71.3 & 78.8 & 46.1 & 61.3 & 54.4 & 65.5 & 39.3 & 40.9 & 61.7 \\
        \midrule 
        ALT$_{g-only}$                  & 63.5 & 63.8 &	94.9 & 68.9 & 74.4 & 84.6 & 39.7 & 61.1 & 49.3 &	68.8 & 43.4 & 50.8 & \textbf{63.6}\\
        ALT$_{RandConv}$                    & 63.6 & 65.8 & 92.5 & 69.1 & 75.1 & 84.5 & 40.1 & 61.7 & 50.8 & 68.4 & 43.4 & 55.2 & \textbf{64.2}\\
        ALT$_{AugMix}$                      & 65.7 & 68.2 & 93.2 & 71.9 & 74.2 & 86.0 & 40.2 & 62.9 & 49.1 & 68.5 & 43.5 & 53.3 & \textbf{64.7}\\
        \bottomrule 
    \end{tabular}
    }
    \smallskip
    \caption[SSDG PACS]{
        Single-source domain generalization accuracy (\%) on PACS~\cite{csurka2017domain}. 
        \textit{X$\rightarrow$Y} implies X is the source and Y is the target dataset.
        \textit{P: photo; A: art-painting; C: cartoon; S: sketch.}
        Performance is reported as mean of 5 repetitions.
        Standard deviation values are in the appendix.
    }
    \label{tab:results_pacs}
\end{table*}

\begin{table*}[t]
    \centering
    \resizebox{0.89\linewidth}{!}{
    \begin{tabular}{@{}l ccc ccc ccc ccc c@{}}
        \toprule
        \textbf{Method} 
        & \textbf{A$\rightarrow$C} & \textbf{A$\rightarrow$P} & \textbf{A$\rightarrow$R} 
        & \textbf{C$\rightarrow$A} & \textbf{C$\rightarrow$P} & \textbf{C$\rightarrow$R} 
        & \textbf{P$\rightarrow$A} & \textbf{P$\rightarrow$C} & \textbf{P$\rightarrow$R} 
        & \textbf{R$\rightarrow$A} & \textbf{R$\rightarrow$C} & \textbf{R$\rightarrow$P} 
        & \textbf{Avg.}\\
        \midrule
        ERM                             & 42.61 & 59.18 & 69.45 & 48.37 & 56.09 & 59.38 & 46.07 & 40.18 & 68.19 & 63.12 & 45.13 & 74.34 & 56.00 \\
        SagNet~\cite{nam2021reducing}   & 42.18 & 56.03 & 67.34 & 46.68 & 53.89 & 57.88 & 45.49 & 40.09 & 67.11 & 61.39 & 48.32 & 72.79 & 54.93\\
        RandConv~\cite{xu2020robust}    & 43.98 & 55.28 & 67.31 & 45.49 & 56.58 & 59.03 & 43.80 & 43.19 & 66.50 & 57.62 & 48.26 & 72.97 & 55.00\\
        AugMix~\cite{hendrycks2019augmix}   & 45.31 & 61.88 & 71.88 & 49.30 & 58.93 & 62.24 & 50.04 & 42.59 & 71.51 & 64.10 & 47.56 & 75.95 & 58.44 \\
        RandConv{+}AugMix               & 42.61 & 54.43 & 65.62 & 43.70 & 55.04 & 57.91 & 43.24 & 41.71 & 65.52 & 59.17 & 48.18 & 71.17 & 53.94 \\
        \midrule 
        ALT$_{g-only}$                  & 47.26 & 61.14 & 71.21 & 48.88	& 57.81 & 60.99	& 48.15 & 46.70 & 69.30 & 64.85 & 52.84 & 76.28 & \textbf{58.78} \\
        ALT$_{RandConv}$                & 48.33 & 61.19 & 71.75 & 50.13 & 58.82 & 62.26 & 49.21 & 47.03 & 70.53 & 64.88 & 53.10 & 76.07 & \textbf{59.44}\\
        ALT$_{AugMix}$                  & 48.06 & 61.16 & 71.12 & 50.43 & 58.84 & 61.84 & 49.32 & 47.55 & 70.64 & 64.86 & 53.27 & 76.29 & \textbf{59.45} \\
        \bottomrule 
    \end{tabular}
    }
    \smallskip
    \caption[SSDG Office-Home]{Single-source domain generalization accuracy (\%) on Office-Home~\cite{venkateswara2017deep}.
    \textit{X$\rightarrow$Y} implies X is the source and Y is the target dataset.
    \textit{R: real; A: art; C: clipart; P: product.}
    Performance is reported as mean of 5 repetitions.
    Standard deviation values are in the appendix.
    }
    \label{tab:results_officehome}
\end{table*}
We validate our approach with extensive empirical analysis of ALT and its constituent parts using three popularly used domain generalization benchmarks.

\medskip
\noindent\textbf{Datasets.}
The SSDG setup is as follows: we train on a single source domain, and evaluate its performance on unobserved target (or test) domains with no access to any data from them during training. We demonstrate the effectiveness of our approach using three popular domain generalization benchmark datasets:
\textbf{(a) \textit{PACS}}~\cite{li2017deeper} consists of images belonging to 7 classes from 4 domains (photo, art painting, cartoon, sketch);
we choose one domain as the source and the rest as target domains.
\textbf{(b) \textit{Office-Home}}~\cite{venkateswara2017deep} consists of images belonging to 65 classes from 4 domains (art, clipart, real, product);
we choose one domain as the source and the rest as target domains.
\textbf{(c) \textit{Digits}}: we follow the setting from Volpi~ \textit{et al.}~\cite{volpi2018generalizing} and use 1000 images from MNIST~\cite{lecun1998mnist} as the source dataset, and USPS~\cite{denker1988neural}, SVHN~\cite{netzer2011reading}, MNIST-M and SYNTH~\cite{ganin2015unsupervised} as the target datasets.

\medskip
\noindent\textbf{Evaluation.}
For all datasets, we train models on each individual domain, and test on the remaining domains. We provide fine-grained results on each test set as well as the average domain generalization performance. 
We compare with several state-of-the-art techniques on SSDG and compare three variants of our methods: ALT$_{g-only}$ refers to the simplest form of our method that only uses the adversary network during training without an explicit diversity module $r$. ALT$_{RandConv}$ and ALT$_{AugMix}$ utilize RandConv and AugMix, respectively, as the diversity module, where the consistency is now placed as explained in Equation \ref{eq:L_KL}.

\subsection{PACS}
\paragraph{Baselines.}
Our baselines are JiGen~\cite{carlucci2019domain}, ADA \cite{volpi2018generalizing}, AugMix~\cite{hendrycks2019augmix}, RandConv~\cite{xu2020robust}, and SagNet~\cite{nam2021reducing} -- designed to reduce style bias using normalization techniques.
We also implement a combination of RandConv and AugMix -- i.e. instead of the ALT formulation of using a diversity module and our adversary network, we use two diversity modules (RandConv and AugMix) and enforce the same consistency as Equation~\ref{eq:L_KL}.
This allows us to compare how effective the adversary network is, compare to using two sources of diversity.
We use ResNet18~\cite{he2016deep} pre-trained on ImageNet as our model architecture and train all models for $2000$ iterations with batch-size of $32$, learning rate $0.004$, \texttt{SGD} optimizer with cosine annealing learning rate scheduler, weight decay of $0.0001$, and momentum $0.9$.
For ALT, we set consistency coefficient $\lambda_{KL}{=}0.75$, adversarial learning rate $\eta_{adv}{=}5e{-}5$, number of adversarial steps $m_{adv}{=}10$ and $w_r{=}1.0$.

\medskip
\noindent\textbf{Results.}
Results are shown in Table~\ref{tab:results_pacs}.
We observe that ALT without a diversity module (ALT$_{g-only}$) surpasses generalization performance of all prior methods including diversity methods RandConv and AugMix and the previous best SagNet~\cite{nam2021reducing}.
ALT with adaptive diversity further improves the results and ALT$_{AugMix}$ establishes a new state-of-the-art accuracy of $64.7\%$.
All three variants of ALT are better than the combination of RandConv+AugMix, providing further evidence that adversarially learned transformations are more effective than combinations of diversity-based augmentations.
The \textit{Sketch (S)} target domain (human drawn black-and-white sketches of real objects) has been the most difficult for previous methods; the difficulty can be observed in terms of performance in columns $A{\rightarrow}S$, $C{\rightarrow}S$, and $P{\rightarrow}S$.
ALT significantly improves the performance on the sketch target domain.
Generalizing from photos as source to C, S, A as targets is a very realistic setting, since large-scale natural image datasets such as ImageNet~\cite{deng2009imagenet} are widely used and publicly available, while data for sketches, cartoons, and paintings are limited.
ALT is the best model under this realistic setting.

\subsection{Office-Home}
\paragraph{Baselines.}
For OfficeHome, we follow the protocol from the previous state-of-the-art Sagnet~\cite{nam2021reducing} and use ResNet50 as the model architecture.
Note that we do not perform any hyperparameter tuning for OfficeHome and directly apply identical training settings and hyperparameters from PACS.

\medskip
\noindent\textbf{Results.}
Table~\ref{tab:results_officehome} shows the results on Office-Home.
We observe that RandConv (previous best on Digits) and SagNet (previous best on PACS) perform worse than ERM on OfficeHome, while AugMix is better by $2.44\%$.
The combination of RandCon+AugMix is also worse than the ERM baseline.
All three variants of ALT surpass prior results, with ALT$_{AugMix}$ resulting in the best accuracy of $59.45\%$.
The most difficult target domain for previous methods is \textit{Clipart (C)}, possibly because most clip-art images have white backgrounds, while real world photos (R) and product images are naturally occurring.
ALT improves performance in each case with C as the target domain.
An observation similar to PACS can also be made here -- ALT is the best model under the realistic setting of generalizing from widely available real photos (R) to other domains.

\begin{table*}[t]
    \centering
    \small
    \resizebox{0.78\linewidth}{!}{
    \begin{tabular}{@{}l lllll l @{}}
        \toprule
        \textbf{Method} & \textbf{MNIST-10K} & \textbf{MNIST-M} & \textbf{SVHN} & \textbf{USPS} & \textbf{SYNTH} & \textbf{Target Avg.} \\ 
        \midrule
        ERM                                 & 98.40 {\footnotesize$\pm$ 0.84} & 58.87 {\footnotesize$\pm$ 3.73 } & 33.41 {\footnotesize$\pm$ 5.28 } & 79.27 {\footnotesize$\pm$ 2.70 } & 42.43 {\footnotesize$\pm$ 5.46 } & 53.50 {\footnotesize$\pm$ 4.23 } \\
        ADA~\cite{volpi2018generalizing}    &  N/A  & 60.41 & 35.51 & 77.26 & 45.32 & 54.62 \\ 
        M-ADA~\cite{qiao2020learning}       & 99.30 & 67.94 & 42.55 & 78.53 & 48.95 & 59.49 \\ 
        ESDA~\cite{volpi2019addressing} & 99.30 {\footnotesize$\pm$ 0.10} & 81.60 {\footnotesize$\pm$ 1.60} & 48.90 {\footnotesize$\pm$ 5.20} & 84.00 {\footnotesize$\pm$ 1.20} & 62.20 {\footnotesize$\pm$ 1.30} & 69.12 {\footnotesize$\pm$ 2.33}\\
        AugMix~\cite{hendrycks2019augmix}   & 98.53 {\footnotesize$\pm$ 0.18} & 53.36 {\footnotesize$\pm$ 1.59} & 25.96 {\footnotesize$\pm$ 0.80} & 96.12 {\footnotesize$\pm$ 0.72} & 42.90 {\footnotesize$\pm$ 0.60} & 54.59 {\footnotesize$\pm$ 0.50}\\
        RandConv~\cite{xu2020robust}        & 98.85 {\footnotesize$\pm$ 0.04 } & 87.76 {\footnotesize$\pm$ 0.83} &	57.62 {\footnotesize$\pm$ 2.09} & 83.36 {\footnotesize$\pm$	0.96} & 62.88 {\footnotesize$\pm$ 0.78} & 72.88 {\footnotesize$\pm$ 0.58} \\
        \midrule 
        ALT$_{g-only}$                      & 98.46 {\footnotesize$\pm$0.27} & 74.28 {\footnotesize$\pm$ 1.36} & 52.25 {\footnotesize$\pm $1.54} & 94.99 {\footnotesize$\pm$ 0.68} & 68.44 {\footnotesize$\pm$ 0.98} & 72.49{\footnotesize$\pm$ 0.87} \\
        ALT$_{RandConv}$                    & 98.46 {\footnotesize$\pm$ 0.25} & 76.90 {\footnotesize$\pm$ 1.42} & 53.78 {\footnotesize$\pm$ 1.97} & 95.40 {\footnotesize$\pm$ 0.72} & 69.40 {\footnotesize$\pm$ 1.07} & \textbf{73.87} {\footnotesize$\pm$ 1.03}\\
        ALT$_{AugMix}$                      & 98.55 {\footnotesize$\pm$ 0.11} & 75.98 {\footnotesize$\pm$ 0.59} & 55.01 {\footnotesize$\pm$ 1.34} & 96.17 {\footnotesize$\pm$ 0.45} & 69.93 {\footnotesize$\pm$ 2.17} & \textbf{74.38} {\footnotesize$\pm$ 0.86}\\
        \bottomrule
    \end{tabular}
    }
    \smallskip
    \caption{Single-source domain generalization accuracy (\%) on digit classification, with MNIST-10K as source and MNIST-M~\cite{ganin2015unsupervised}, SVHN~\cite{netzer2011reading}, USPS~\cite{denker1988neural}, and SYNTH~\cite{ganin2015unsupervised} as target domains. 
    Note: ADA and M-ADA do not report standard deviation.}
    \label{tab:results_digits}
\end{table*}
\subsection{Digits}
\paragraph{Baselines.}
Our baselines include a na\"ive ``source-only'' model trained using expected risk minimization (ERM) on the source dataset, M-ADA~\cite{qiao2020learning} -- an adversarial data augmentation method, and AugMix~\cite{hendrycks2019augmix} 
and RandConv~\cite{xu2020robust} which exploit diversity through consistency constraints.
We also compare with ESDA~\cite{volpi2019addressing}, an evolution-based search procedure over a pre-defined set of augmentations~\cite{cubuk2019autoaugment}.
We use DigitNet~\cite{volpi2018generalizing} as the model architecture for all models for a fair comparison.
All models are trained for $T{=}10000$ iterations, with batch-size of $32$, learning rate of $0.0001$, using the \texttt{Adam} optimizer.
For ALT, we set the consistency coefficient $\lambda_{KL}{=}0.75$, adversarial learning rate $\eta_{adv}{=}5e{-}6$, number of adversarial steps $m_{adv}{=}10$, and equal weight $w_r{=}1.0$ for diversity and adversary networks.

\medskip
\noindent\textbf{Results.}
Table~\ref{tab:results_digits} shows that pixel-level adversarial training approaches (ADA and M-ADA) offer only marginal improvements over the na\"ive ERM baseline. 
The results for diversity-promoting data augmentation methods are mixed -- while AugMix is only $1.09\%$ better than ERM, RandConv provides a significant boost. Interestingly, the base version of our approach, ALT$_{g-only}$, which is exclusively based on adversarial training, is significantly better than pixel-level adversarial training.
More importantly, it is also better than diversity method AugMix, while performing lower than RandConv by a small margin $0.39\%$.
When we trained ALT with adaptive diversity (ALT$_{RandConv}$ and ALT$_{AugMix}$), we achieved the best performance, beating previous state-of-the-art.
SVHN and SYNTH are the hardest target domains as they contain real-world images of street signs or house number signs, whereas USPS is closely correlated with MNIST, both being black-and-white centered images of handwritten digits, and MNIST-M is derived from MNIST but with different backgrounds. 
AugMix fares poorly on both real-world datasets, but is able to generalize well to MNIST-M and USPS.
Although AugMix results in an average accuracy of $54.59\%$ on the target domains, when used in conjunction with ALT, the ALT$_{AugMix}$ leads to a large gain of $19.79\%$, highlighting the significance of the adversary network.
\begin{figure*}[t]
    \centering
    \small
    \raisebox{-0.5\height}{\includegraphics[width=0.445\linewidth]{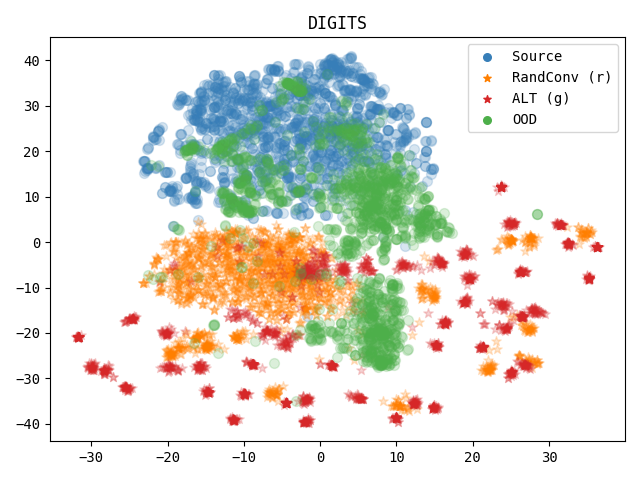}}
    \raisebox{-0.5\height}{\includegraphics[width=0.445\linewidth]{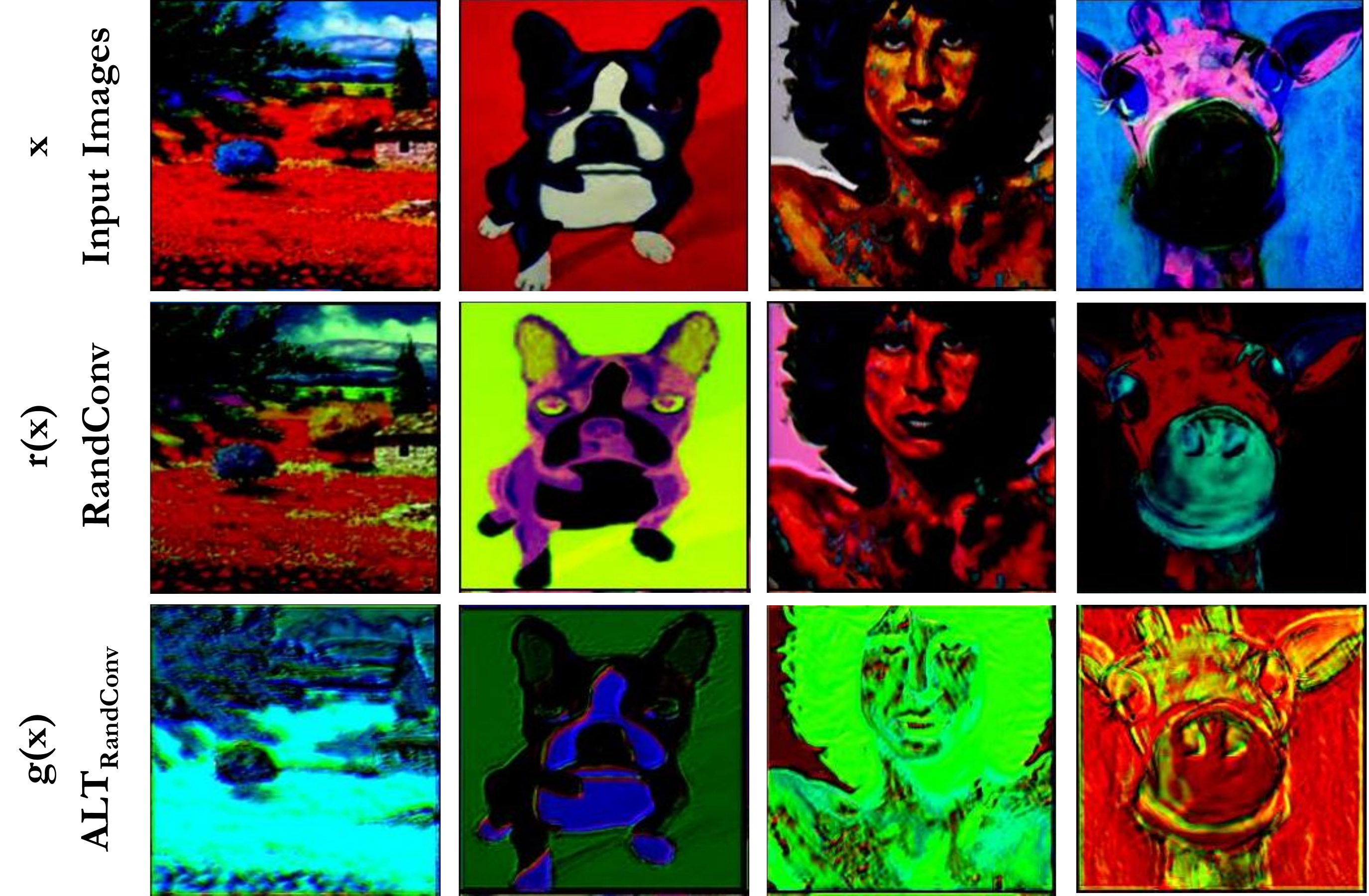}}
    \caption{
        \textit{(Left)} tSNE plot showing the discrepancy between the source distribution (MNIST) and the out-of-distribution datasets for the ``Digits'' benchmark.
        The diversity introduced by ALT is much larger and wide-spread than data augmentation techniques such as RandConv.
        \textit{(Right)} Qualitative Comparison of PACS images transformed by RandConv data augmentation vs. ALT (ALT$_{RandConv}$), illustrating the wide range of transformations learned by ALT.
        }
    \label{fig:viz_augs}
\end{figure*}

\begin{figure*}[t]
    \centering
    \includegraphics[width=0.9\linewidth]{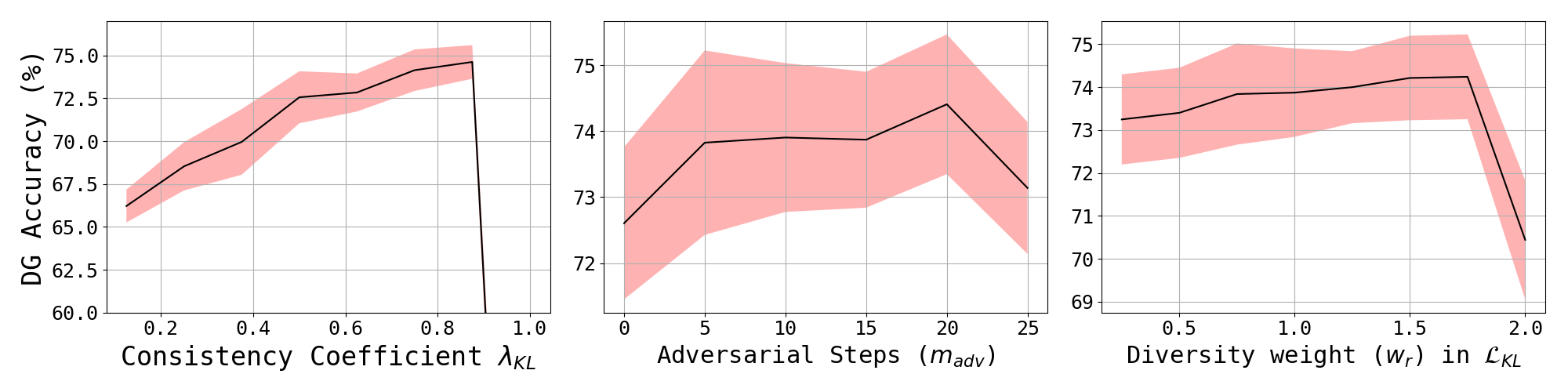}
    \caption{\textbf{Analysis:} We study the effect of each hyper-parameter in ALT on the average accuracy using the Digits benchmark (shown as 1 standard deviation around the mean over 5 runs).
    We observe that the consistency \textit{(left)} is generally important until a certain point, after which it becomes harmful; \textit{(middle)} taking more adversarial steps improves performance;
    \textit{(right)} surprisingly, we find that the trade-off between diversity and adversity is non- trivial and dataset dependent.
    In our benchmarking experiments (Tables~\ref{tab:results_pacs}, \ref{tab:results_officehome} , \ref{tab:results_digits}) we do not perform any hyper-parameter tuning, and set $w_r{=}1$, i.e. equal weight to adversity and diversity.
    }
    \label{fig:digits_hp}
\end{figure*}

\section{Analysis of ALT}
In this section we study the various components of ALT, and provide insights into their impact on generalization.

\subsection{ALT is better than na\"ive diversity.}

Our first big insight is that ALT without an explicit diversity (ALT$_{g-only}$) module still outperforms all the top performing methods across the benchmarks we evaluated on, indicating that learned adversarial transformations are a powerful way to train classifiers for generalization. 

Our next observation is that ALT makes the choice of diversity module fairly arbitrary. We see this effect on multiple benchmarks -- for example, on the Digits benchmark shown in Table~\ref{tab:results_digits}, AugMix has a relatively poor generalization performance when compared with the baseline ERM whereas ALT$_{Augmix}$ achieves state of the art. This is again seen in the Office-Home benchmark shown in Table~\ref{tab:results_officehome}, where RandConv is worse than ERM, but ALT$_{RandConv}$ is the best performing method. 
Thus, irrespective of the choice of diversity module, the adversarially learned transformations benefit generalization on all benchmarks.

In Figure~\ref{fig:viz_augs} \textit{(left panel)} we analyze the diversity introduced by ALT on the Digits benchmark, in comparison to the source distribution the target (OOD) distribution and the distribution of RandConv augmentations.
While RandConv does simulate a domain shift compared to the source, most RandConv points are clustered close to each other.
However, the diversity due to ALT is considerably larger and ALT samples are spread widely across the tSNE space.
We believe this is because data augmentation functions have a fixed types of diversity (random convolution filter in the case of RandConv), while ALT \textit{searches} for adversarial transformations for each batch -- this leads to novel types of diversity for each batch of training samples.
We also show qualitative examples of the image transformations learned with ALT in Figure~\ref{fig:viz_augs}, and it is clear that ALT achieves far more diverse and larger transformations of the input images than previous data augmentation techniques.
A similar comparison of ALT with AugMix~\cite{hendrycks2019augmix} is shown in the supplementary material.

\subsection{Effect of Varying ALT Hyperparameters.}
The three main hyper-parameters that control ALT are: 
    \textbf{(1)} $\lambda_{KL}$ -- the coefficient in Eq.~\ref{eq:L_KL} which decides the weight for the KL-divergence consistency in the total loss, 
    \textbf{(2)} $m_{adv}$ -- the number of adversarial steps in the adversarial maximization of Eq.~\ref{eq:adv_max}, and 
    \textbf{(3)} $w_r$ -- the diversity weight which controls the interaction between the diversity module $r()$ and the adversary network $g()$ in Eq.~\ref{eq:p_mix}.
We investigate the effect of each of these on domain generalization accuracy in Figure~\ref{fig:digits_hp}.
The first plot shows that the consistency coefficient $\lambda_{KL}$ is impactful and a higher value leads to better generalization.
However at $\lambda_{KL}=1.0$ the accuracy degenerates to random performance; this is expected as the classifier loss gets $1-\lambda_{KL}{=}0$ weight.
From the second plot, we observe that the optimal number of adversarial steps is around 20.
Note that performance at all non-zero values of $m_{adv}$ that we tried $(5, 10, 15, 20, 25)$ is greater than previous state-of-the-art.
The importance of the adversarial module is evident from the third plot -- performance at $w_r=0$ (adversarial module only) is higher than performance of $w_r=2$ (diversity module only), and the combination of both modules yields the best result.
Clearly, the adversarial component is a critical factor that causes improvements in generalization.

\section{Conclusion}
In this paper, we address the problem of single source domain generalization. 
Our approach, Adversarially Learned Transformations (ALT) updates a convolutional network to learn plausible image transformations of the source domain that can fool the classifier during training; and enforces a consistency constraint on the predictions on clean images and transformed images.
ALT is a significant improvement over prior methods that utilize pixel-wise perturbations.
We showed that this strategy outperforms all existing techniques, including standard data augmentation methods, on multiple benchmarks because it is able to generate a diverse set of large transformations of the source domain. 
We also find that ALT can be naturally combined with existing diversity modules like RandConv or AugMix to improve their performance.
We studied components of ALT through extensive ablations and analysis to obtain insights into its performance gains.
Our studies indicate that na\"ive diversity alone is insufficient, but needs to be combined with adversarially learned transformations to maximize generalization performance. 

\medskip
\noindent\textbf{Acknowledgements:}
This work was performed under the auspices of the U.S. Department of Energy by the Lawrence Livermore National Laboratory under Contract No. DE-AC52-07NA27344, Lawrence Livermore National Security, LLC. and was supported by the LDRD Program under project 22-ERD-006 with IM release number LLNL-JRNL-836221. 
BK's efforts were supported by 22-DR-009.
TG, CB, and YY were supported by NSF RI grants  \#1816039 and \#2132724.

{\small
\bibliographystyle{ieee_fullname}
\bibliography{tgokhale}
}

\appendix
\section*{Appendix}

This appendix contains training settings, additional results and visualizations to supplement the main paper.
We also discuss the limitations of ALT and scope for future work in this direction.
Code to reproduce experiments has been released publicly: \url{https://github.com/tejas-gokhale/ALT}.

\section{Training Settings} 
Table~\ref{tab:hyperparams} shows the training settings and hyperparameters used for experiments on each benchmark. 
See Algorithm 1 in the main paper for context and relevant equations.

\section{Detailed Results}
We provide detailed results including standard deviation values for our models on the PACS and Office-Home benchmarks, for each source domain.
We compare these with RandConv~\cite{xu2020robust},  AugMix~\cite{hendrycks2019augmix}, and a combination of RandConv and AugMix which utilizes AugMix as one of the two augmentations in the consistency constraint of RandConv.
Results of the PACS experiments are shown in Tables \ref{tab:results_pacs_P}, \ref{tab:results_pacs_A}, \ref{tab:results_pacs_C}, \ref{tab:results_pacs_S}, when using P, A, C, and S as the source datasets, respectively.
Results of the Office-Home experiments are show in Tables \ref{tab:results_oh_R}, \ref{tab:results_oh_A}, \ref{tab:results_oh_C}, \ref{tab:results_oh_P}, when using R, A, C, and P as the source datasets, respectively.

\section{Visualizations}
\begin{figure*}
    \centering
    \includegraphics[width=\linewidth]{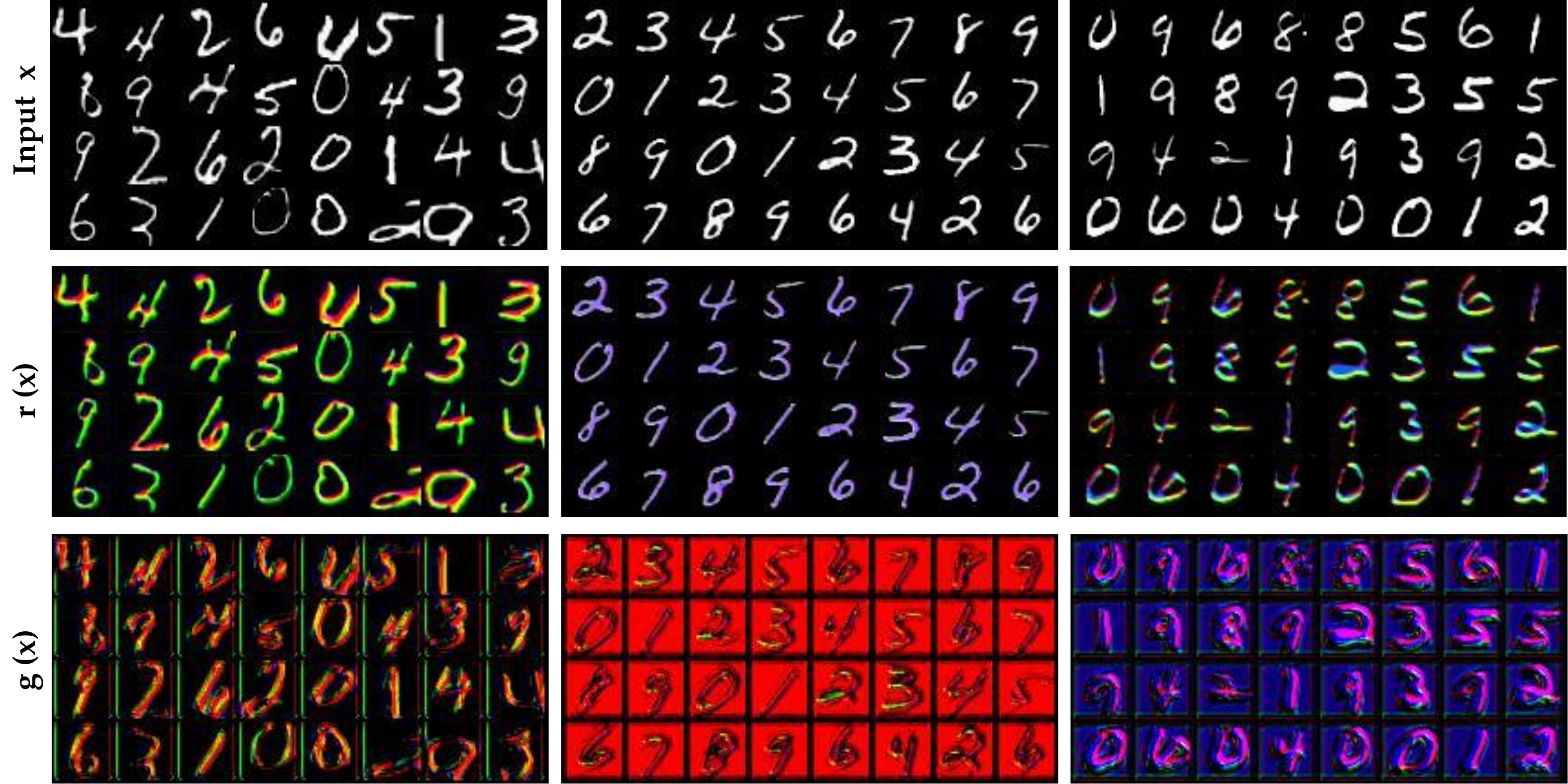}
    \caption{Digits: Comparison of images transformed by RandConv and ALT$_{RandConv}$ with MNIST10k as source dataset.}
    \label{fig:supp_digits}
\end{figure*}

\begin{figure*}
    \centering
    \includegraphics[width=\linewidth]{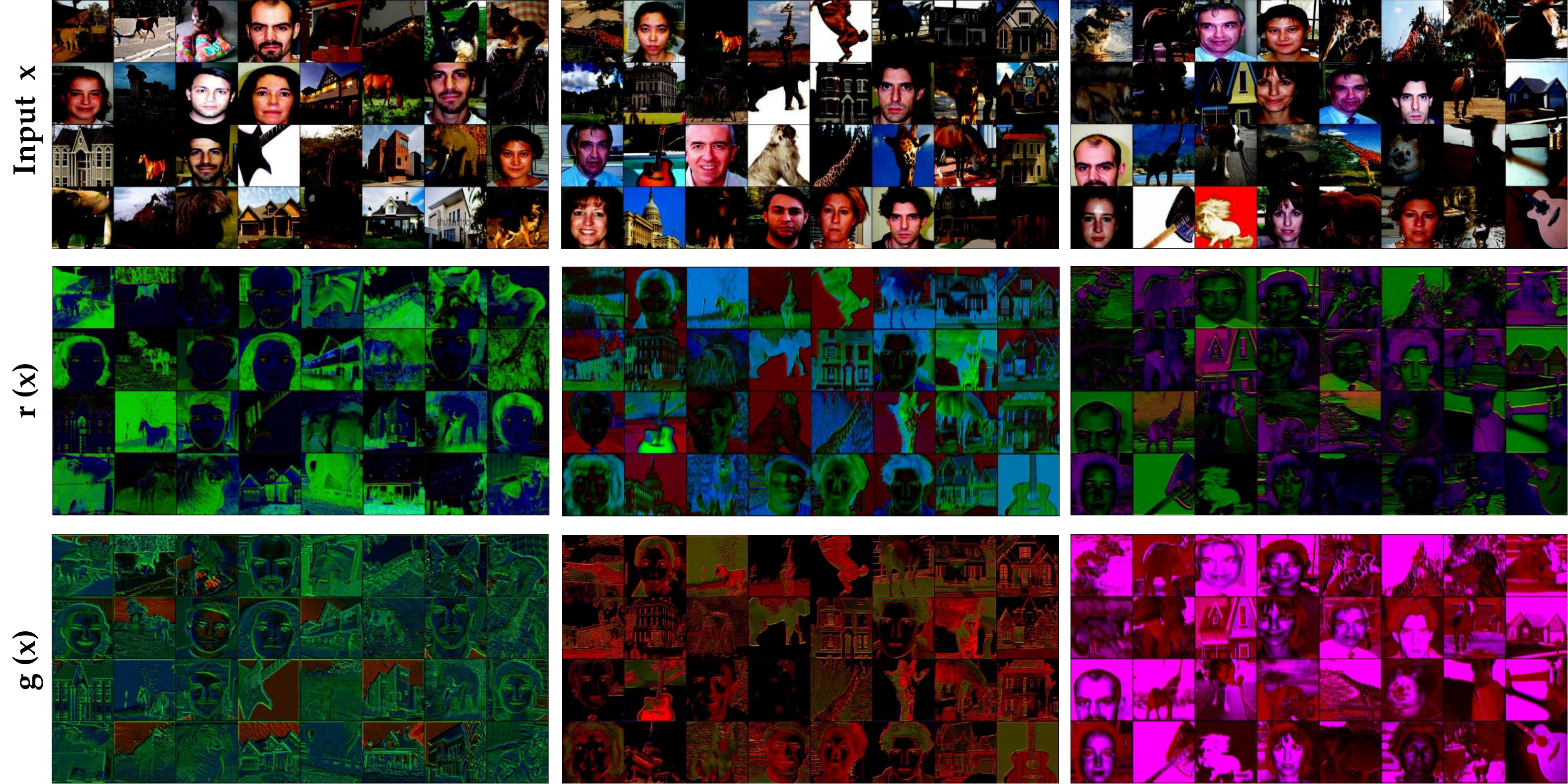}
    \caption{PACS: Comparison of images transformed by RandConv and ALT$_{RandConv}$ with \textit{Photo} as source dataset.}
    \label{fig:supp_pacs_p}
\end{figure*}

\begin{figure*}
    \centering
    \includegraphics[width=\linewidth]{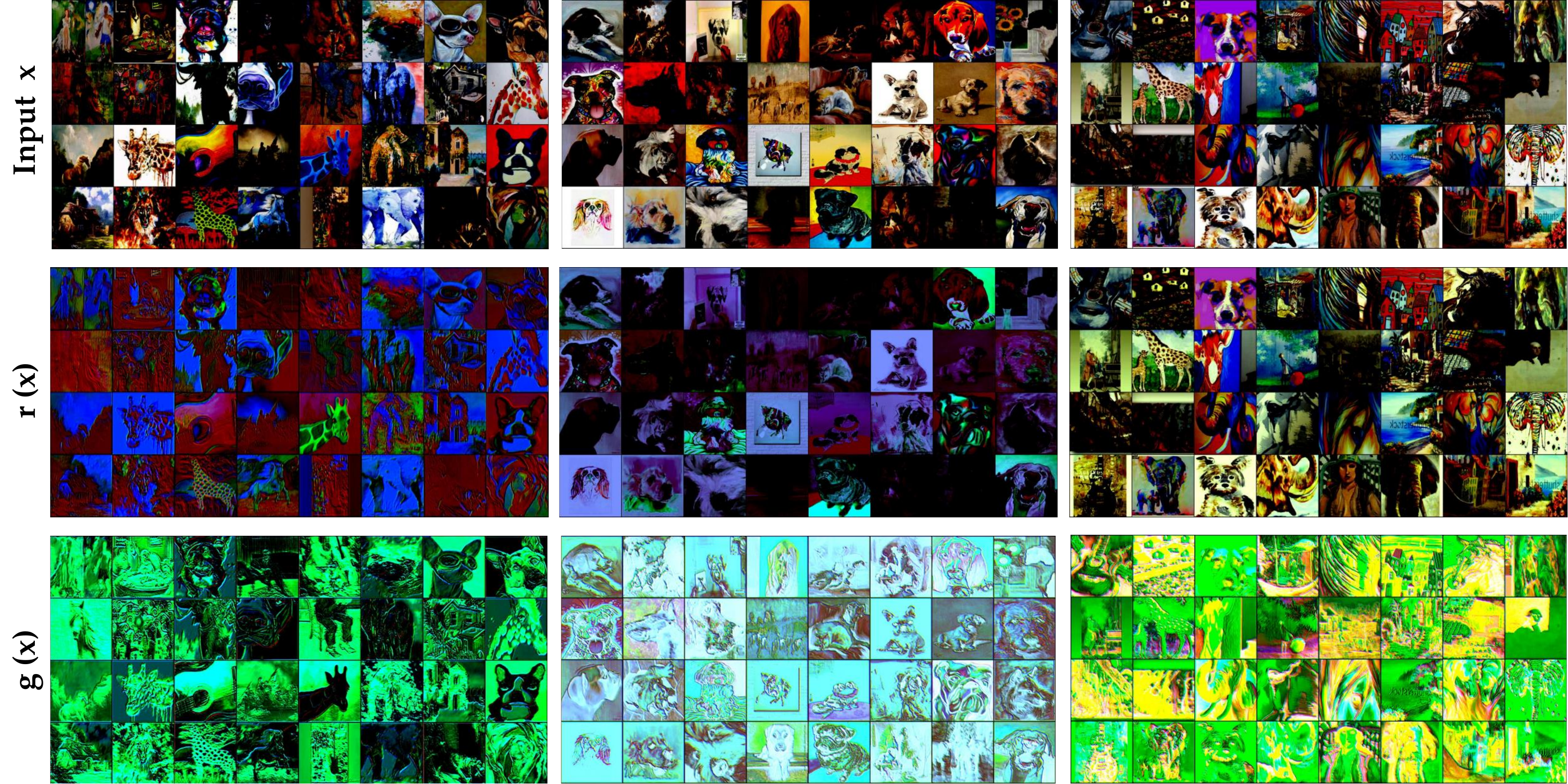}
    \caption{PACS: Comparison of images transformed by RandConv and ALT$_{RandConv}$ with \textit{Art-Painting} as source dataset.}
    \label{fig:supp_pacs_a}
\end{figure*}

\begin{figure*}
    \centering
    \includegraphics[width=\linewidth]{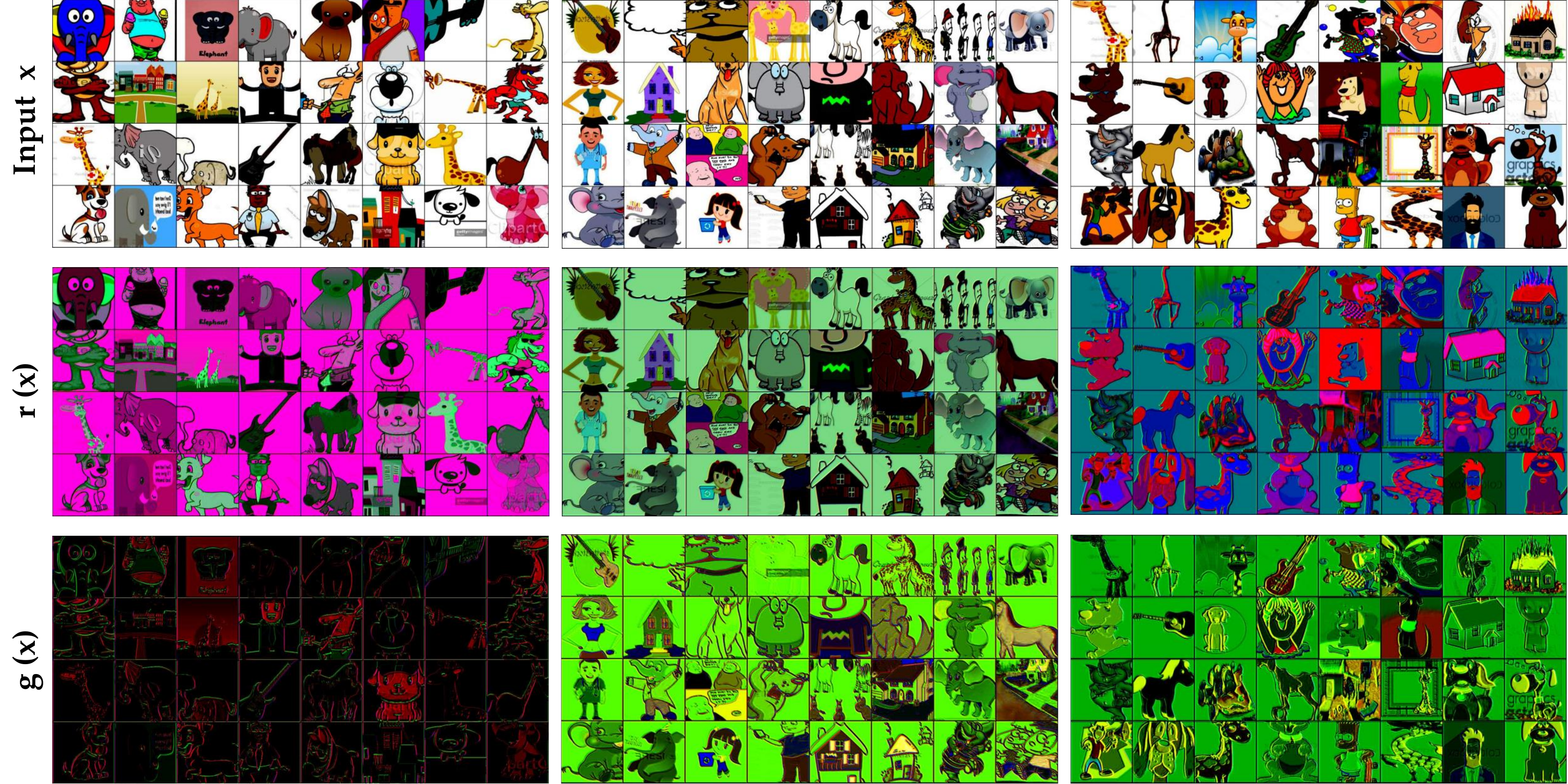}
    \caption{PACS: Comparison of images transformed by RandConv and ALT$_{RandConv}$ with \textit{Cartoon} as source dataset.}
    \label{fig:supp_pacs_c}
\end{figure*}

\begin{figure*}
    \centering
    \includegraphics[width=\linewidth]{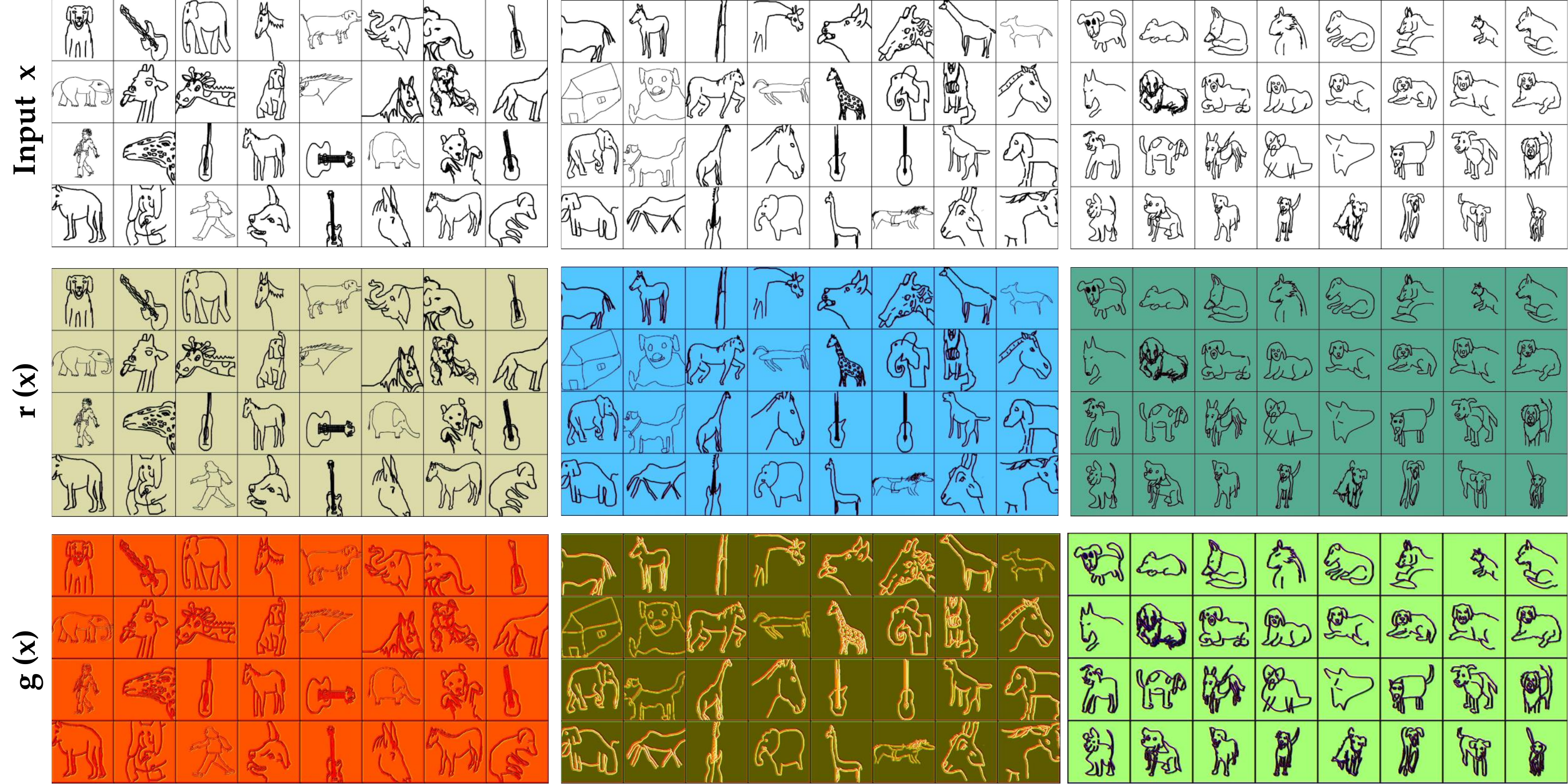}
    \caption{PACS: Comparison of images transformed by RandConv and ALT$_{RandConv}$ with \textit{Sketch} as source dataset.}
    \label{fig:supp_pacs_s}
\end{figure*}

\begin{figure*}
    \centering
    \includegraphics[width=\linewidth]{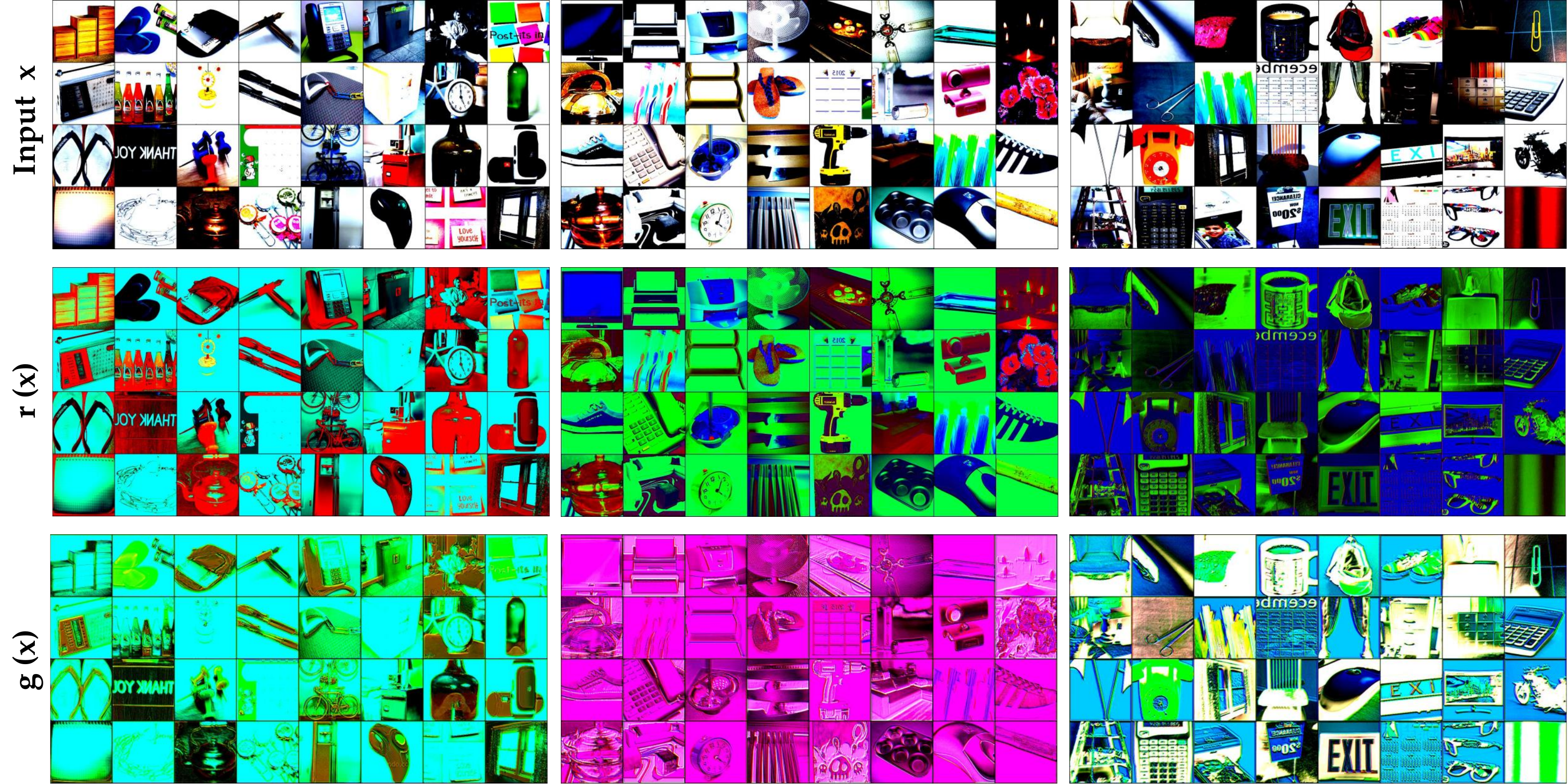}
    \caption{Office-Home: Comparison of images transformed by RandConv and ALT$_{RandConv}$ with \textit{Real} as source dataset.}
    \label{fig:supp_oh_r}
\end{figure*}

\begin{figure*}
    \centering
    \includegraphics[width=\linewidth]{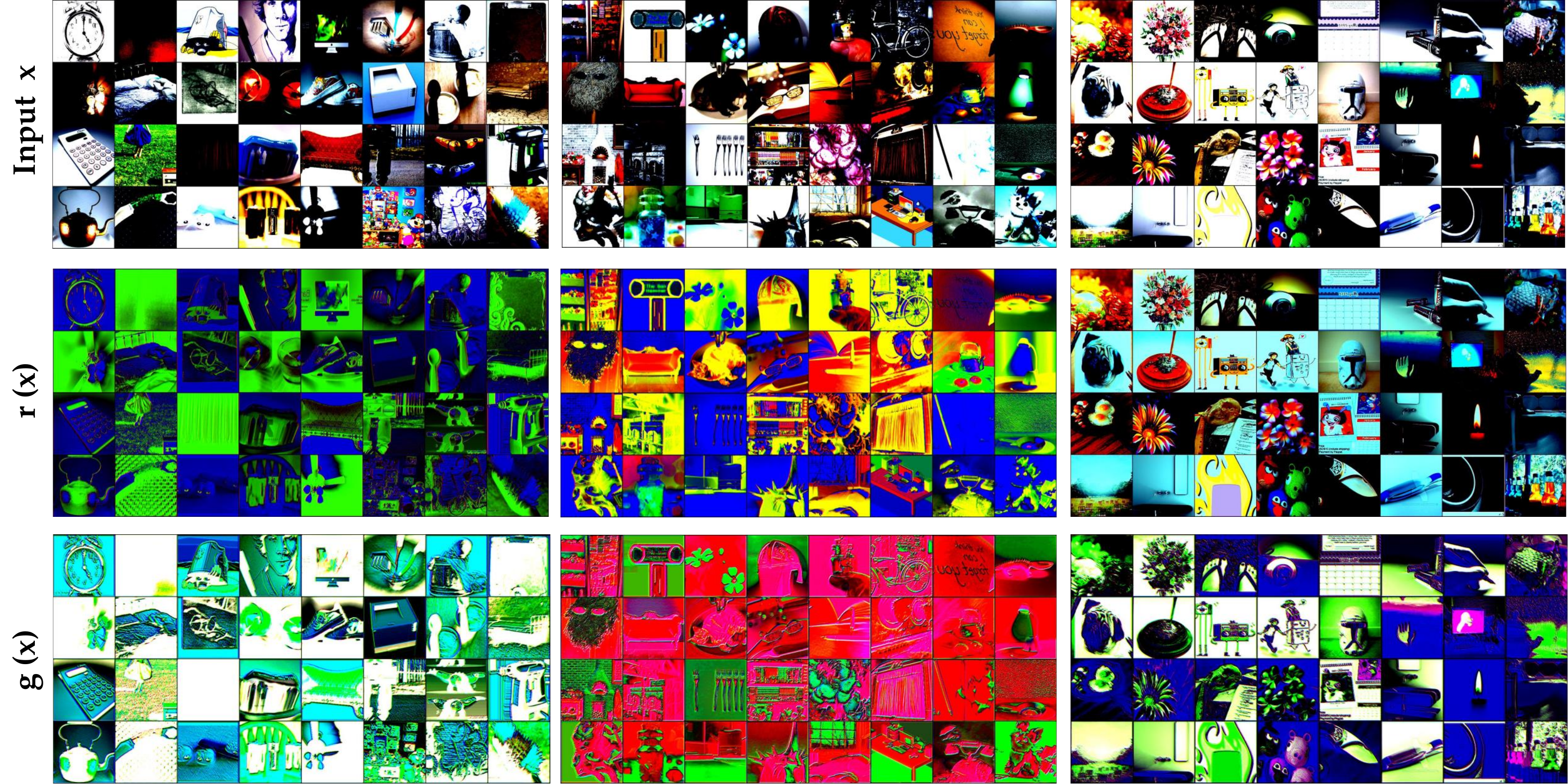}
    \caption{Office-Home: Comparison of images transformed by RandConv and ALT$_{RandConv}$ with \textit{Art} as source dataset.}
    \label{fig:supp_oh_a}
\end{figure*}

\begin{figure*}
    \centering
    \includegraphics[width=\linewidth]{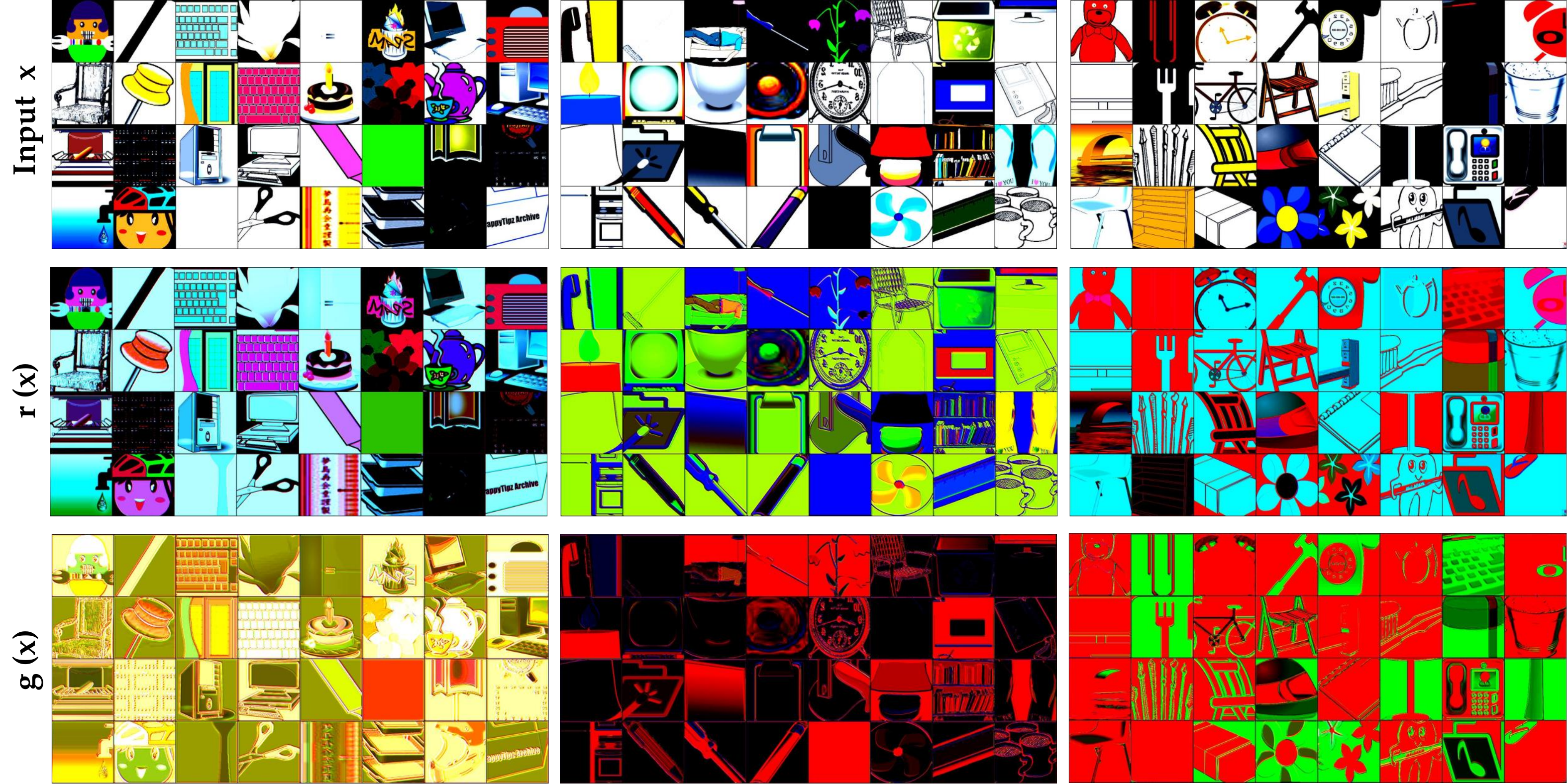}
    \caption{Office-Home: Comparison of images transformed by RandConv and ALT$_{RandConv}$ with \textit{Clipart} as source dataset.}
    \label{fig:supp_oh_c}
\end{figure*}

\begin{figure*}
    \centering
    \includegraphics[width=\linewidth]{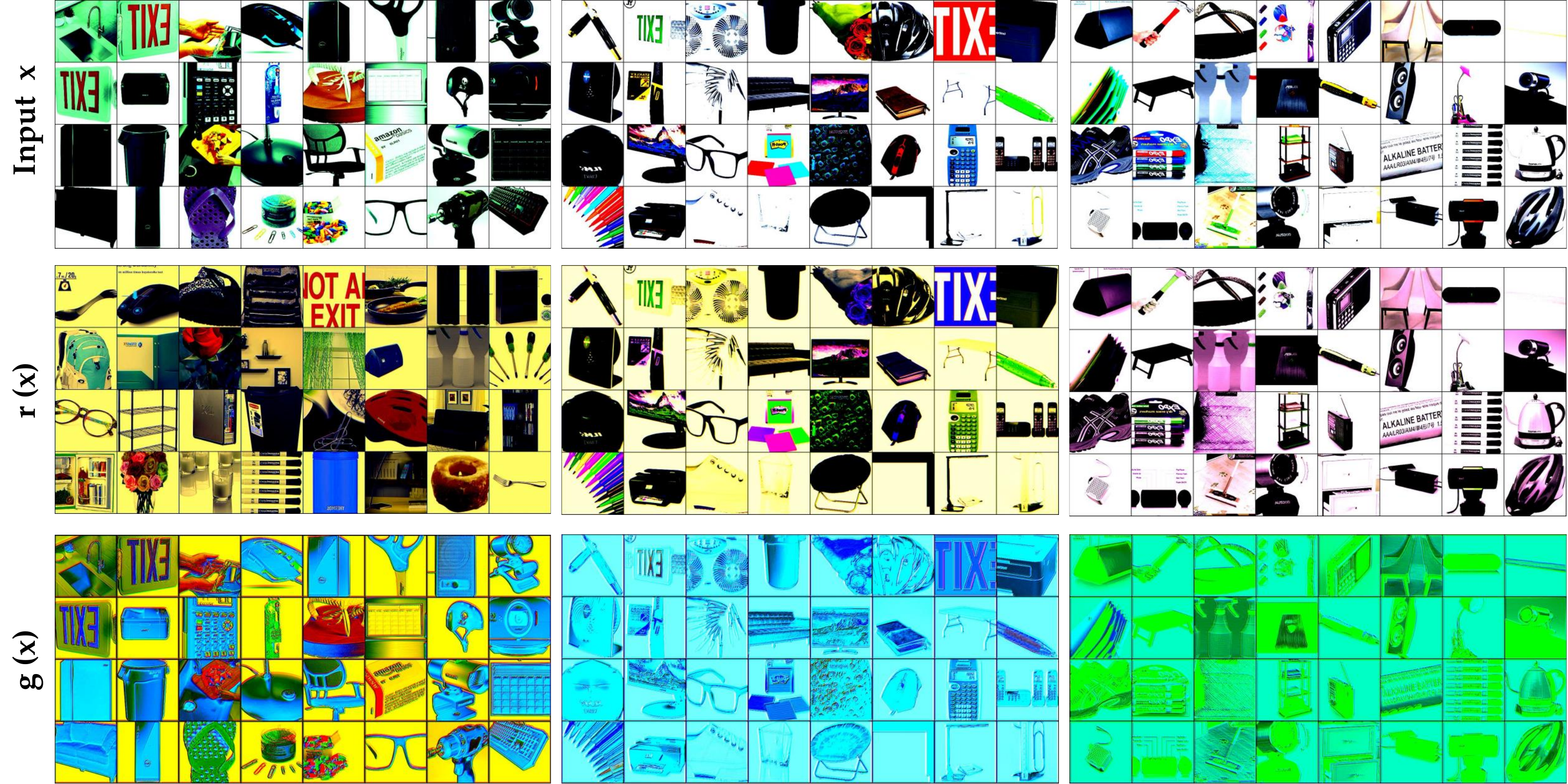}
    \caption{Office-Home: Comparison of images transformed by RandConv and ALT$_{RandConv}$ with \textit{Product} as source dataset.}
    \label{fig:supp_oh_p}
\end{figure*}

In this section, we provide additional visualizations and qualitative examples for augmented images generated by ALT, for Digits (Figure~\ref{fig:supp_digits}), PACS (Figures~\ref{fig:supp_pacs_p}, \ref{fig:supp_pacs_a}, \ref{fig:supp_pacs_c}, \ref{fig:supp_pacs_s}), and Office-Home PACS (Figures~\ref{fig:supp_oh_r}, \ref{fig:supp_oh_a}, \ref{fig:supp_oh_c}, \ref{fig:supp_oh_p}).
In each figure, the first row shows input images $\xx$, the second row shows the outputs of the diversity module $r(\xx)$, and the third row shows the outputs of the adversary network $g(\xx)$.

\begin{figure*}
    \centering
    \includegraphics[height=2.5in]{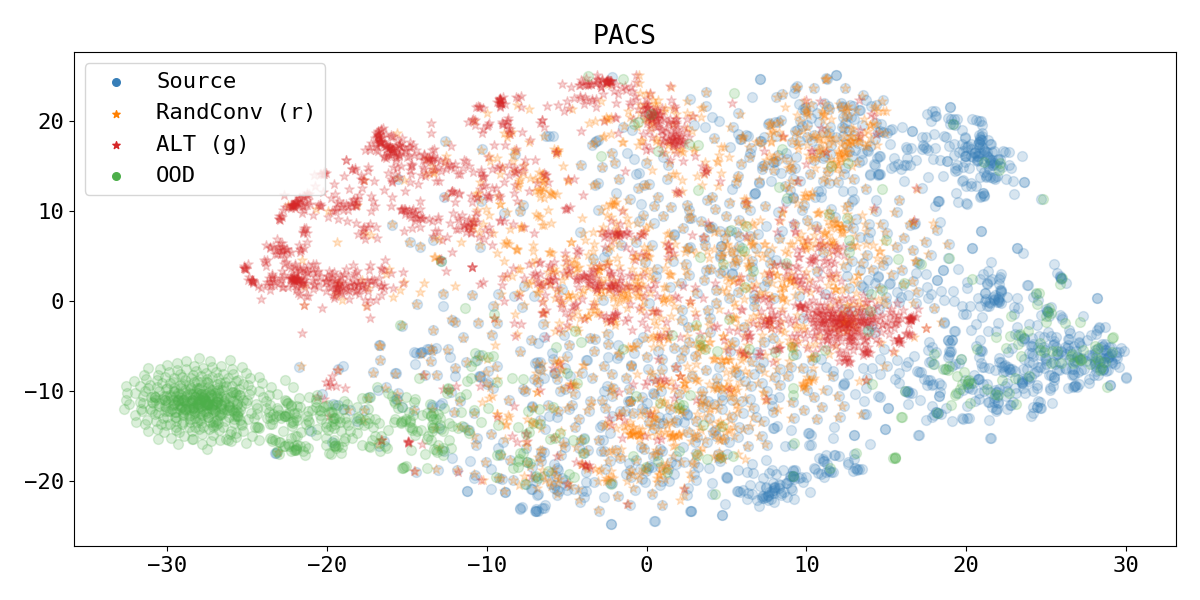}
    \caption{tSNE plot showing the discrepancy between the source distribution and the out-of-distribution datasets for the PACS benchmark.}
    \label{fig:viz_tsne_pacs}
\end{figure*}

\begin{figure*}
    \centering
    \includegraphics[height=2.5in]{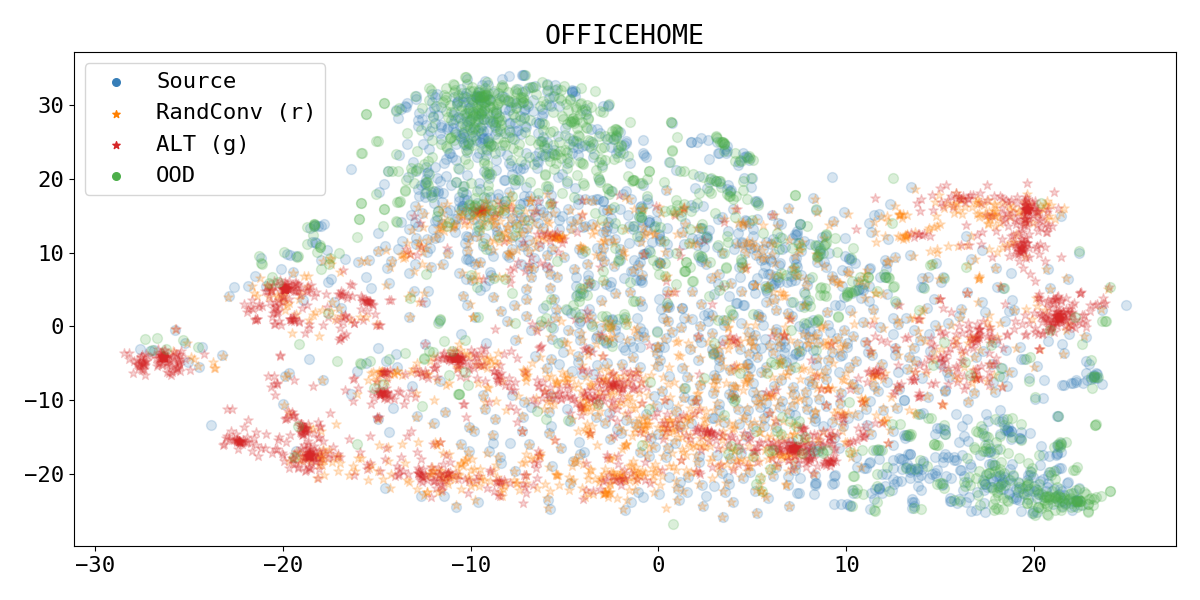}
    \caption{tSNE plot showing the discrepancy between the source distribution and the out-of-distribution datasets for the OfficeHome benchmark.}
    \label{fig:viz_tsne_officehome}
\end{figure*}

In Figure~\ref{fig:viz_tsne_pacs} and \ref{fig:viz_tsne_officehome} we show an illustration of the diversity introduced by ALT in comparison to the source distribution, the target (OOD) distribution, and the distribution of RandConv augmentations, for the PACS and OfficeHome benchmarks respectively.
The diversity introduced by ALT is much larger and wide-spread than data augmentation techniques such as RandConv.

\begin{table}
    \centering
    \resizebox{\linewidth}{!}{
    \begin{tabular}{@{}l ccc@{}} 
        \toprule
        \textbf{Variable} & \textbf{Digits} & \textbf{PACS} & \textbf{Office-Home} \\
        \midrule
        $f$ architecture    & DigitNet~\cite{volpi2018generalizing} & ResNet18~\cite{he2016deep} & ResNet50~\cite{he2016deep}\\
        $g$ architecture    & \multicolumn{3}{c}{
            \vtop{
                \hbox{\strut \{conv$_{kernel=3, stride=1, padding=1, leakyReLU_{p=0.2}}\}\times4$}
                }
            }\\
        $\rho_0, \phi_0$    & \multicolumn{3}{c}{Kaiming Normal Initialization~\cite{he2015delving}}\\ 
        $T$                 & 10000 & 2000  & 2000  \\
        $T_{pre}$           & 1250  & 400   & 400   \\
        $\eta$              & 1e-4  & 0.004 & 0.004 \\
        $m_{adv}$           & 10    & 10    & 10 \\
        $\eta_{adv}$        & 5e-6  & 5e-5  & 5e-5 \\
        $w_r$               & 1.0   & 1.0   & 1.0 \\ 
        $\lambda_{KL}$      & 0.75  & 0.75  & 0.75 \\
        \bottomrule
    \end{tabular}
    }
    \smallskip
    \caption{Training settings and hyper-parameters for experiments on each benchmark.}
    \label{tab:hyperparams}
\end{table}
\section{Limitations and Future Directions} 
In this paper we have explored the effectiveness of ALT on three standard domain generalization benchmarks.
For fair comparison, for each baseline model, we use the same model architecture and training settings as the backbone -- for instance, ERM, RandConv, AugMix and ALT are all trained with the same hyperparameters as shown in Table~\ref{tab:hyperparams}.
For significance of results, we have repeated each experiment (including those in analyses) for 5 different seeds and have reported mean and standard deviation.

\subsection{Complexity of Adversary Network.}
One limitation (and therefore scope for future work) is that we have considered one family of architecture for our adversary network $g$ -- fully convolutional image-to-image translation networks. 
We conduct additional analysis to understand how this choice affects generalization performance, and compare performance when using between 2 and 6 convolutional layers.
We reuse all other training settings from our benchmark model $ALT_{RandConv}$ on both Digits and PACS.
Results are shown in Table~\ref{tab:gvars}.

For PACS and OfficHome, we observe that all ALT models compared are better than previous baselines including AugMix and RandConv.
For Digits, we observe that performance of ALT with a 2-layer $g$ is close to RandConv, and is greater than all previous baselines for higher depth of the network.
We do not see a clear correlation across datasets between the number of layers and the domain generalization performance.
Investigating the dynamics of model capacity of the adversary network and how it may affect domain generalization, is an interesting direction for future work.

\begin{table}[t]
    \centering
    \small
    \begin{tabular}{@{}cc ccccc@{}}
        \toprule
         & \hphantom & \multicolumn{5}{c}{\textbf{FCN Number of Layers}}\\
        \textbf{Benchmark} && 2 & 3 & 4 & 5 & 6 \\
        \midrule 
        Digits      &&72.75~~&73.74~~&74.10~~&73.87~~&74.15 \\
        PACS        &&63.40~~&63.92~~&64.41~~&64.20~~&63.78 \\
        OfficeHome  
        &&59.67~~&59.56~~&59.79~~&59.42~~&59.45 \\
        \bottomrule
    \end{tabular}
    \smallskip
    \caption{Effect of the depth (number of convolutional layers) of the adversity network $g$ on average domain generalization on all three benchmarks.}
    \label{tab:gvars}
\end{table}

It may be possible that more complex generative architectures (i.e.~greater complexity of transformations) may be needed for larger domain shift is larger, to model diversity and adversity for a given source domain.
Thus the choice of architecture for $g$ is an interesting direction; nevertheless, in this paper we show that the simple fully convolutional architecture gives us performance boosts in all three datasets.

We believe that ideas presented in this paper, although evaluated on image classification, have the potential of being widely applicable to many other vision tasks for domain generalization.
They may also be applied to other application areas such as audio or text, where the transformation function $g$ may take different forms.

\begin{table*}[!b]
    \centering
    \small
    \begin{tabular}{@{}l llll ll@{}}
        \toprule
        \textbf{Method} & \textbf{Photo$^\star$}    & \textbf{Art-Painting} & \textbf{Cartoon}  & \textbf{Sketch}       & \textbf{Target Avg.}  & \textbf{PACS Avg.} \\
        \midrule
        RandConv           & 96.407	\tiny{$\pm$0.757} &	61.309	\tiny{$\pm$2.316} &	37.577	\tiny{$\pm$2.257} &	50.463	\tiny{$\pm$9.018} &	49.783	\tiny{$\pm$4.255} &	61.439	\tiny{$\pm$3.217}\\
        AugMix             & \textbf{99.532}	\tiny{$\pm$0.438} &	68.633	\tiny{$\pm$0.950} &	33.788	\tiny{$\pm$1.205} &	36.304	\tiny{$\pm$2.801} &	46.242	\tiny{$\pm$1.122} &	59.564	\tiny{$\pm$0.930}\\
        RandConv + AugMix  & 98.363	\tiny{$\pm$0.438} &	65.527	\tiny{$\pm$3.060} &	39.300 \tiny{$\pm$6.237} & 40.901	\tiny{$\pm$5.073} & 48.576 \tiny{$\pm$4.031} & 61.023	\tiny{$\pm$3.001} \\
        \midrule
        ALT$_{g-only}$ & 99.064	\tiny{$\pm$ 0.286} & \textbf{68.770}	\tiny{$\pm$ 0.932} & 43.387	\tiny{$\pm$ 1.142} & 50.832	\tiny{$\pm$ 2.937} & 54.330	\tiny{$\pm$ 1.078} & 65.513	\tiny{$\pm$ 0.757} \\
        ALT$_{RandConv}$ & 98.947 \tiny{$\pm$0.234} & 68.740	\tiny{$\pm$0.702} &	40.828	\tiny{$\pm$2.537} &	\textbf{56.024}	\tiny{$\pm$2.009} &	\textbf{55.197}	\tiny{$\pm$0.498} & 66.135	\tiny{$\pm$0.330}\\
        ALT$_{AugMix}$ & 99.298	\tiny{$\pm$0.438} &	68.506	\tiny{$\pm$0.836} &	\textbf{43.507}	\tiny{$\pm$2.615} &	53.271	\tiny{$\pm$4.149} &	55.094	\tiny{$\pm$1.876} &	\textbf{66.145} \tiny{$\pm$1.387}\\
        \bottomrule
    \end{tabular}
    \smallskip 
    \caption{SSDG performance on PACS for the P$\rightarrow$ACS setting. $^\star$Source Domain. \textbf{bold}: best result. }
    \label{tab:results_pacs_P}
\end{table*}
\begin{table*}[!b]
    \centering
    \small
    \begin{tabular}{@{}l llll ll@{}}
        \toprule
        \textbf{Method} & \textbf{Photo}      & \textbf{Art Painting$^\star$} & \textbf{Cartoon}  & \textbf{Sketch}       & \textbf{Target Avg.}  & \textbf{PACS Avg.} \\
        \midrule
        RandConv         & 87.281	\tiny{$\pm$0.796} &	85.437	\tiny{$\pm$0.532} &	61.143	\tiny{$\pm$2.752} &	60.519	\tiny{$\pm$4.050} &	69.648	\tiny{$\pm$2.152} &	73.595	\tiny{$\pm$1.582} \\
        AugMix          & \textbf{95.317} \tiny{$\pm$0.422} & \textbf{93.077}	\tiny{$\pm$1.276} &	64.061	\tiny{$\pm$0.361} &	55.027	\tiny{$\pm$2.195} &	71.469	\tiny{$\pm$0.637} &	76.871	\tiny{$\pm$0.581}\\
        RandConv + AugMix   & 90.743 \tiny{$\pm$0.781} & 90.481	\tiny{$\pm$0.638} &	64.206	\tiny{$\pm$2.238} &	62.515	\tiny{$\pm$2.854} &	72.488	\tiny{$\pm$1.731} & 76.986	\tiny{$\pm$1.177} \\
        \midrule
        ALT$_{g-only}$ & 94.934 \tiny{$\pm$0.269} & 91.058 \tiny{$\pm$0.720} & 63.524 \tiny{$\pm$1.821} & 63.813 \tiny{$\pm$2.249} & 74.090 \tiny{$\pm$1.086} & 78.332	\tiny{$\pm$0.845}\\
        ALT$_{RandConv}$ & 93.593	\tiny{$\pm$0.328} &	92.596	\tiny{$\pm$1.036} &	64.044	\tiny{$\pm$0.635} &	65.991	\tiny{$\pm$1.130} & 74.543	\tiny{$\pm$0.537} & 79.056	\tiny{$\pm$0.609}\\
        ALT$_{AugMix}$ & 93.174	\tiny{$\pm$0.437} &	91.442	\tiny{$\pm$0.638} &	\textbf{65.683}	\tiny{$\pm$1.656} &	\textbf{68.226}	\tiny{$\pm$2.453} &	\textbf{75.694}	\tiny{$\pm$1.214} &	\textbf{79.631}	\tiny{$\pm$0.856}\\
        \bottomrule
    \end{tabular}
    \smallskip 
    \caption{SSDG performance on PACS for the A$\rightarrow$PCS setting. $^\star$Source Domain. \textbf{bold}: best result. }
    \label{tab:results_pacs_A}
\end{table*}
\begin{table*}[!b]
    \centering
    \small
    \begin{tabular}{@{}l llll ll@{}}
        \toprule
        \textbf{Method} & \textbf{Photo}    & \textbf{Art-Painting} & \textbf{Cartoon$^\star$}  & \textbf{Sketch}       & \textbf{Target Avg.}  & \textbf{PACS Avg.} \\
        \midrule
        RandConv           & 73.677	\tiny{$\pm$1.814} &	57.051	\tiny{$\pm$1.764} &	91.66	\tiny{$\pm$0.876} &	72.855	\tiny{$\pm$2.314} &	67.861	\tiny{$\pm$1.550} &	73.810	\tiny{$\pm$1.317} \\
        AugMix              & 84.599 \tiny{$\pm$0.997} & 68.281	\tiny{$\pm$2.085} &	\textbf{96.287}	\tiny{$\pm$0.940} & 71.097	\tiny{$\pm$0.609} &	74.659	\tiny{$\pm$1.088} &	80.066	\tiny{$\pm$0.897}\\
        RandConv + AugMix  & 78.790	\tiny{$\pm$0.975} & 65.400	\tiny{$\pm$1.611} &	93.840	\tiny{$\pm$1.020} &	71.285	\tiny{$\pm$2.730} & 71.825	\tiny{$\pm$1.315} &	77.329	\tiny{$\pm$1.105}\\
        \midrule
        ALT$_{g-only}$ & 84.575	\tiny{$\pm$1.047} &	68.867	\tiny{$\pm$2.126} &	94.768	\tiny{$\pm$0.43} &	74.421	\tiny{$\pm$0.441} &	75.954	\tiny{$\pm$1.119} &	80.658	\tiny{$\pm$0.929}\\
        ALT$_{RandConv}$ & 83.916	\tiny{$\pm$0.51} &	68.086	\tiny{$\pm$1.901} &	95.190	\tiny{$\pm$0.686} &	\textbf{74.487}	\tiny{$\pm$0.505} &	75.496	\tiny{$\pm$0.799} &	80.420	\tiny{$\pm$0.644} \\
        ALT$_{AugMix}$ & \textbf{85.964}	\tiny{$\pm$1.098} &	\textbf{71.943}	\tiny{$\pm$1.234} &	94.599	\tiny{$\pm$0.560} &	74.172	\tiny{$\pm$0.752} &	\textbf{77.360}	\tiny{$\pm$0.734} &	\textbf{81.670}	\tiny{$\pm$0.667}\\
        \bottomrule
    \end{tabular}
    \smallskip 
    \caption{SSDG performance on PACS for the C$\rightarrow$PAS setting. $^\star$Source Domain. \textbf{bold}: best result. }
    \label{tab:results_pacs_C}
\end{table*}
\begin{table*}
    \centering\small
    \begin{tabular}{@{}l llll ll@{}}
        \toprule
        \textbf{Method} & \textbf{Photo}    & \textbf{Art-Painting} & \textbf{Cartoon}  & \textbf{Sketch$^\star$}       & \textbf{Target Avg.}  & \textbf{PACS Avg.} \\
        \midrule
        RandConv           & 46.132	\tiny{$\pm$4.879} &	\textbf{52.168}	\tiny{$\pm$1.623} &	\textbf{63.942}	\tiny{$\pm$2.219} &	94.264	\tiny{$\pm$0.673} &	\textbf{54.081}	\tiny{$\pm$1.959} &	\textbf{64.126}	\tiny{$\pm$1.465}\\
        AugMix          & 46.731 \tiny{$\pm$2.916} & 37.852	\tiny{$\pm$1.878} &	58.575	\tiny{$\pm$1.747} &	94.221	\tiny{$\pm$0.711} &	47.719	\tiny{$\pm$1.723} &	59.345	\tiny{$\pm$1.268}\\
        RandConv + AugMix  & \textbf{54.359}	\tiny{$\pm$0.819} &	46.074	\tiny{$\pm$2.709} &	61.246	\tiny{$\pm$1.245} &	94.171	\tiny{$\pm$0.582} &	53.893	\tiny{$\pm$0.945} &	63.963	\tiny{$\pm$0.787}\\
        \midrule
        ALT$_{g-only}$ & 49.305	\tiny{$\pm$2.775} &	39.658	\tiny{$\pm$3.423} &	61.109	\tiny{$\pm$1.853} &	94.573	\tiny{$\pm$0.466} &	50.024	\tiny{$\pm$2.408} &	61.161	\tiny{$\pm$1.726}\\
        ALT$_{RandConv}$ & 51.305 \tiny{$\pm$0.866} & 41.787	\tiny{$\pm$1.174} &	62.773	\tiny{$\pm$1.089} &	\textbf{94.724}	\tiny{$\pm$0.527} &	51.955	\tiny{$\pm$0.791} &	62.647	\tiny{$\pm$0.571}\\
        ALT$_{AugMix}$  & 49.078 \tiny{$\pm$2.072} & 40.186	\tiny{$\pm$2.494} &	62.901	\tiny{$\pm$0.358} &	94.271	\tiny{$\pm$0.624} &	50.721	\tiny{$\pm$1.414} &	61.609	\tiny{$\pm$1.103}\\
        \bottomrule
    \end{tabular}
    \smallskip 
    \caption{SSDG performance on PACS for the S$\rightarrow$PAC setting. $^\star$Source Domain. \textbf{bold}: best result. }
    \label{tab:results_pacs_S}
\end{table*}
\clearpage
\begin{table*}
    \centering\small
    \begin{tabular}{@{}l llll ll@{}}
        \toprule
        \textbf{Method} & \textbf{Real$^\star$}    & \textbf{Art} & \textbf{Clipart}  & \textbf{Product}       & \textbf{Target Avg.}  & \textbf{Office-Home Avg.} \\
        \midrule
        RandConv            & 83.028 \tiny{$\pm$2.067} & 59.021	\tiny{$\pm$0.916} &	47.269	\tiny{$\pm$1.251} &	72.172	\tiny{$\pm$0.418} &	59.487	\tiny{$\pm$0.792} &	65.372	\tiny{$\pm$1.096}\\
        AugMix              & 87.294 \tiny{$\pm$1.21} &	64.101	\tiny{$\pm$0.882} &	47.564	\tiny{$\pm$0.158} &	75.956	\tiny{$\pm$0.32} &	62.54	\tiny{$\pm$0.345} &	68.729	\tiny{$\pm$0.490}\\
        RandConv + AugMix   & 81.514 \tiny{$\pm$0.515} & 59.167	\tiny{$\pm$0.722} &	48.180	\tiny{$\pm$1.024} &	71.166	\tiny{$\pm$0.445} &	59.504	\tiny{$\pm$0.256} &	65.007	\tiny{$\pm$0.226}\\
        \midrule 
        ALT$_{g-only}$      & 86.514 \tiny{$\pm$0.622} & 64.622	\tiny{$\pm$0.490} &	\textbf{53.327}	\tiny{$\pm$0.344} &	76.276	\tiny{$\pm$0.117} &	64.742	\tiny{$\pm$0.122} &	70.185	\tiny{$\pm$0.138}\\
        ALT$_{RandConv}$    & \textbf{87.477} \tiny{$\pm$1.042} & \textbf{64.879}	\tiny{$\pm$0.439} &	53.097	\tiny{$\pm$0.554} &	76.066	\tiny{$\pm$0.447} &	64.681	\tiny{$\pm$0.290} &	\textbf{70.380}	\tiny{$\pm$0.312}\\
        ALT$_{AugMix}$      & 86.560 \tiny{$\pm$0.980} & 64.860	\tiny{$\pm$0.267} &	53.271	\tiny{$\pm$0.799} &	\textbf{76.286}	\tiny{$\pm$0.347} &	\textbf{64.806}	\tiny{$\pm$0.327} &	70.244	\tiny{$\pm$0.461}\\
        \bottomrule
    \end{tabular}
    \smallskip 
    \caption{SSDG performance on Office-Home for the R$\rightarrow$ACP setting. $^\star$Source Domain. \textbf{bold}: best result. }
    \label{tab:results_oh_R}
\end{table*}
\begin{table*}
    \centering\small
    \begin{tabular}{@{}l llll ll@{}}
        \toprule
        \textbf{Method} & \textbf{Real}    & \textbf{Art$^\star$} & \textbf{Clipart}  & \textbf{Product}       & \textbf{Target Avg.}  & \textbf{Office-Home Avg.} \\
        \midrule 
        RandConv            & 66.915 \tiny{$\pm$1.069} & 72.428	\tiny{$\pm$2.066} & 42.387	\tiny{$\pm$1.405} &	55.045	\tiny{$\pm$1.547} &	54.782	\tiny{$\pm$1.297} &	59.194	\tiny{$\pm$1.427}\\
        AugMix              & \textbf{71.887} \tiny{$\pm$0.432} & \textbf{80.494}	\tiny{$\pm$1.342} &	45.314	\tiny{$\pm$0.768} &	\textbf{61.882}	\tiny{$\pm$0.382} &	59.694	\tiny{$\pm$0.427} &	\textbf{64.894}	\tiny{$\pm$0.369}\\
        RandConv + AugMix   & 65.620 \tiny{$\pm$0.632} & 71.852	\tiny{$\pm$1.758} &	42.606	\tiny{$\pm$1.026} &	54.434	\tiny{$\pm$0.774} &	54.220	\tiny{$\pm$0.617} &	58.628	\tiny{$\pm$0.862}\\
        \midrule 
        ALT$_{g-only}$      & 71.193 \tiny{$\pm$0.308} & 78.930	\tiny{$\pm$1.146} & 47.340	\tiny{$\pm$0.331} &	61.151	\tiny{$\pm$0.561} &	59.895	\tiny{$\pm$0.283} & 64.654	\tiny{$\pm$0.259}\\
        ALT$_{RandConv}$    & 71.754 \tiny{$\pm$0.286} & 78.025	\tiny{$\pm$1.181} &	\textbf{48.328}	\tiny{$\pm$0.787} &	61.186	\tiny{$\pm$0.429} &	\textbf{60.423}	\tiny{$\pm$0.280} &	64.823	\tiny{$\pm$0.474}\\
        ALT$_{AugMix}$      & 71.122 \tiny{$\pm$0.540} & 79.095	\tiny{$\pm$1.634} &	48.058	\tiny{$\pm$0.632} &	61.156	\tiny{$\pm$0.813} &	60.112	\tiny{$\pm$0.590} & 64.858	\tiny{$\pm$0.518}\\
        \bottomrule
    \end{tabular}
    \smallskip 
    \caption{SSDG performance on Office-Home for the A$\rightarrow$RCP setting. $^\star$Source Domain. \textbf{bold}: best result. }
    \label{tab:results_oh_A}
\end{table*}
\begin{table*}
    \centering\small
    \begin{tabular}{@{}l llll ll@{}}
        \toprule
        \textbf{Method} & \textbf{Real}    & \textbf{Art} & \textbf{Clipart$^\star$}  & \textbf{Product}       & \textbf{Target Avg.}  & \textbf{Office-Home Avg.} \\
        \midrule 
        RandConv            & 58.944 \tiny{$\pm$0.521} & 44.741	\tiny{$\pm$0.714} & 80.320	\tiny{$\pm$1.073} &	56.211	\tiny{$\pm$1.141} &	53.299	\tiny{$\pm$0.711} &	60.054	\tiny{$\pm$0.771}\\
        AugMix              & 62.244 \tiny{$\pm$0.526} & 49.309	\tiny{$\pm$0.879} &	\textbf{81.510}	\tiny{$\pm$0.885} &	\textbf{58.939}	\tiny{$\pm$0.584} &	56.831	\tiny{$\pm$0.530} &	\textbf{63.000}	\tiny{$\pm$0.454}\\
        RandConv + AugMix   & 57.914 \tiny{$\pm$0.730} & 43.698	\tiny{$\pm$0.511} &	77.986	\tiny{$\pm$1.087} &	55.040	\tiny{$\pm$0.683} &	52.217	\tiny{$\pm$0.249} & 58.660	\tiny{$\pm$0.417}\\
        \midrule 
        ALT$_{g-only}$      & 61.968 \tiny{$\pm$0.849} & 49.977	\tiny{$\pm$0.987} &	80.320	\tiny{$\pm$1.073} &	58.779	\tiny{$\pm$0.743} &	56.908	\tiny{$\pm$0.808} &	62.761	\tiny{$\pm$0.733}\\
        ALT$_{RandConv}$    & \textbf{62.264} \tiny{$\pm$0.560} & 50.133	\tiny{$\pm$0.956} &	80.732	\tiny{$\pm$0.637} &	58.819	\tiny{$\pm$0.558} &	\textbf{57.072}	\tiny{$\pm$0.539} &	62.987	\tiny{$\pm$0.455}\\
        ALT$_{AugMix}$      & 61.841 \tiny{$\pm$0.382} & \textbf{50.426}	\tiny{$\pm$1.070} & 80.824	\tiny{$\pm$0.510} &	58.839	\tiny{$\pm$0.559} & 57.035	\tiny{$\pm$0.580} & 62.982	\tiny{$\pm$0.526}\\
        \bottomrule
    \end{tabular}
    \smallskip 
    \caption{SSDG performance on Office-Home for the C$\rightarrow$RAP setting. $^\star$Source Domain. \textbf{bold}: best result. }
    \label{tab:results_oh_C}
\end{table*}

\begin{table*}
    \centering\small
    \begin{tabular}{@{}l llll ll@{}}
        \toprule
        \textbf{Method} & \textbf{Real}    & \textbf{Art} & \textbf{Clipart}  & \textbf{Product$^\star$}       & \textbf{Target Avg.}  & \textbf{Office-Home Avg.} \\
        \midrule 
        RandConv            & 66.318 \tiny{$\pm$0.240} &	43.524	\tiny{$\pm$0.664} &	43.365	\tiny{$\pm$1.058} & 90.135	\tiny{$\pm$0.643} &	51.069	\tiny{$\pm$0.607} &	60.836	\tiny{$\pm$0.372}\\
        AugMix              & \textbf{71.515} \tiny{$\pm$0.706} & \textbf{50.041}	\tiny{$\pm$0.688} &	42.596	\tiny{$\pm$0.619} &	\textbf{91.622}	\tiny{$\pm$0.263} &	54.717	\tiny{$\pm$0.518} & 63.943	\tiny{$\pm$0.453}\\
        RandConv + AugMix   & 65.523 \tiny{$\pm$0.753} & 43.240	\tiny{$\pm$1.454} &	41.710 \tiny{$\pm$0.621} &	89.459	\tiny{$\pm$0.785} &	50.158	\tiny{$\pm$0.900} &	59.983	\tiny{$\pm$0.865}\\
        \midrule 
        ALT$_{g-only}$      & 70.082 \tiny{$\pm$0.532} & 48.842	\tiny{$\pm$0.648} &	46.877	\tiny{$\pm$0.552} &	91.306	\tiny{$\pm$0.544} &	55.267	\tiny{$\pm$0.302} &	64.277	\tiny{$\pm$0.171}\\
        ALT$_{RandConv}$    & 70.530 \tiny{$\pm$0.359} & 49.208	\tiny{$\pm$0.418} &	47.025	\tiny{$\pm$0.498} &	91.577	\tiny{$\pm$0.506} &	55.588	\tiny{$\pm$0.300} &	64.585	\tiny{$\pm$0.212}\\
        ALT$_{AugMix}$      & 70.637 \tiny{$\pm$0.301} & 49.318	\tiny{$\pm$1.008} &	\textbf{47.554}	\tiny{$\pm$0.458} &	91.396	\tiny{$\pm$0.798} &	\textbf{55.837}	\tiny{$\pm$0.383} &	\textbf{64.726}	\tiny{$\pm$0.361}\\
        \bottomrule
    \end{tabular}
    \smallskip 
    \caption{SSDG performance on Office-Home for the P$\rightarrow$RAC setting. $^\star$Source Domain. \textbf{bold}: best result. }
    \label{tab:results_oh_P}
\end{table*}

\begin{figure*}
    \centering
    \includegraphics[height=2.5in]{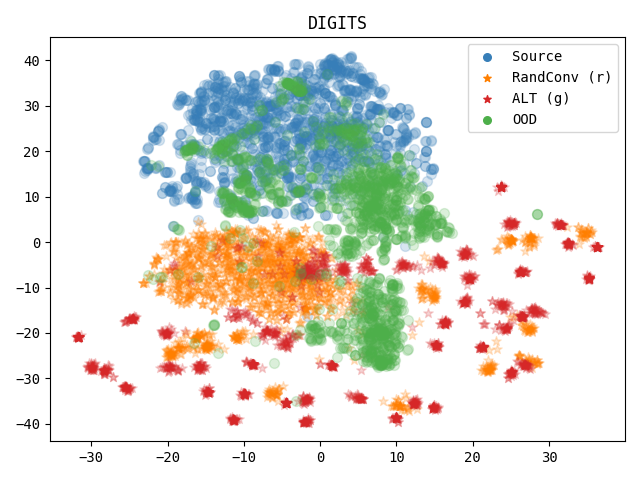}
    \caption{tSNE plot showing the discrepancy between the source distribution and the out-of-distribution datasets for the Digits benchmark.}
    \label{fig:viz_tsne_digits}
\end{figure*}



\end{document}